\documentclass[journal]{IEEEtran}

\ifCLASSINFOpdf

\else
 
\fi

\usepackage{array}
\usepackage{amsmath}
\usepackage{graphicx}
\usepackage{upgreek}
\usepackage{multirow}
\usepackage{color}
\usepackage{pifont}
\usepackage{url}
\usepackage{bm}
\usepackage{threeparttable}
\usepackage{amssymb}
\usepackage{float}
\usepackage[colorlinks,linkcolor=blue]{hyperref}
\usepackage{booktabs} 
\usepackage{amsfonts}
\usepackage{soul}
\soulregister{\cite}7 
\soulregister{\citep}7 
\soulregister{\citet}7 
\soulregister{\ref}7 
\soulregister{\pageref}7 
\usepackage{xcolor}
\usepackage{float}         
\usepackage{algorithm}     
\usepackage{algorithmic} 

\begin{document}

\title {Bridging Sensor Gaps via Attention Gated Tuning for Hyperspectral Image Classification}

\author{Xizhe~Xue,~Haokui~Zhang*,~Haizhao Jing,~Lijie Tao,~Zongwen~Bai,~and~Ying~Li*
\thanks{

This work was completed during X. Xue's Ph.D. period. It was supported in part by the National Natural Science Foundation of China (62271400, 62401471, 62266045) and in part by
2024 Gusu Innovation and Entrepreneurship Leading Talents Program under Grant ZXL2024333. * represents the corresponding author. 

X. Xue, H. Zhang, H. Jing, L. Tao and Y. Li are with Northwestern Polytechnical University, Xi'an, China. (email: xuexizhe@mail.nwpu.edu.cn, hkzhang@nwpu.edu.cn, haizhao\_jing@mail.nwpu.edu.cn, taolijie11111@mail.nwpu.edu.cn lybyp@nwpu.edu.cn). 
Z. Bai is with Yan’an University, Yan’an, China. (e-mail: ydbzw@yau.edu.cn).}

}

\markboth{IEEE Journal of Selected Topics in Applied Earth Observations and Remote Sensing, VOL. xx, xxxx}%
{Shell \MakeLowercase{\textit{et al.}}: Bare Demo of IEEEtran.cls for IEEE Journals}

\maketitle

\begin{abstract}

Data-driven HSI classification methods require high-quality labeled HSIs, which are often costly to obtain. This characteristic limits the performance potential of data-driven methods when dealing with limited annotated samples. Bridging the domain gap between data acquired from different sensors allows us to utilize abundant labeled data across sensors to break this bottleneck. In this paper, we propose a novel Attention-Gated Tuning (AGT) strategy and a triplet-structured transformer model, Tri-Former, to address this issue. The AGT strategy serves as a bridge, allowing us to leverage existing labeled HSI datasets, even RGB datasets to enhance the performance on new HSI datasets with limited samples. Instead of inserting additional parameters inside the basic model, we train a lightweight auxiliary branch that takes intermediate features as input from the basic model and makes predictions. The proposed AGT resolves conflicts between heterogeneous and even cross-modal data by suppressing the disturbing information and enhances the useful information through a soft gate. Additionally, we introduce Tri-Former, a triplet-structured transformer with a spectral-spatial separation design that enhances parameter utilization and computational efficiency, enabling easier and flexible fine-tuning. 
Comparison experiments conducted on three representative HSI datasets captured by different sensors demonstrate the proposed Tri-Former achieves better performance compared to several state-of-the-art methods. Homologous, heterologous and cross-modal tuning experiments verified the effectiveness of the proposed AGT. Code has been released at: \href{https://github.com/Cecilia-xue/AGT}{https://github.com/Cecilia-xue/AGT}.

\end{abstract}

\section{Introduction}
Hyperpectral image (HSI) contains rich spectral information about the composition and properties of objects, resulting in a wide range of applications in earth observation, e.g., mineral exploration~\cite{carrino2018hyperspectral}, plant stress detection~\cite{behmann2014detection}, and environmental science~\cite{transon2018survey}, etc.  Within these applications, HSI classification technology identifies each pixel in an image and assign it to a specific land-cover category. Over the past decade, deep learning (DL) based methods~\cite{lin2013spectral,chen2014deep, chen2015spectral,zhang2017spectral,li2017spectral, chen2016deep,zhang20213} dominate the field of HSI classification. From 2013 until now, various optimized convolutional neural networks (ConvNets) have been proposed for HSI classification~\cite{zhang2017spectral,li2017spectral, chen2016deep}. Among them, 3D ConvNet-based HSI classification methods have become the mainstream in this field, because the 3D convolution is inherently well-suited for processing the HSI with 3D structure. Recently, the emergence of the vision transformer (ViT) has disrupted this situation and researchers have begun exploring how to apply ViT models to handle the HSI classification task. Hong \textit{et al.}~\cite{hong2021spectralformer} were the first to apply pure transformers to HSI classification. In the same trend, impressive results have been achieved by combining the transformer with ConvNet~\cite{sun2022spectral}, architecture search (NAS) algorithm~\cite{3danas} and new absolute position embedding~\cite{gst}. 

The performance of aforementioned data-hungry methods are closely related to the quantity and quality of labeled data. However,  annotating HSIs requires field surveys by professionals, making the process time-consuming and expensive. The number of labeled hyperspectral data is relatively small. To alleviate this problem, some early work~\cite{2019Hyperspectral, yang2017learning} explores how to use additional labelled data from other domains to further push back these limits.  Two-CNN-transfer~\cite{yang2017learning} first investigates transfer learning between the data from the same sensor in HSI classification. On similar trend, method~\cite{2019Hyperspectral} concentrates on  exploiting data from other sensors and transferring the knowledge. Such previous work~\cite{2019Hyperspectral, yang2017learning} both bridge the domain gap between source and target domains by fine tuning the ConvNets. However, such fine tuning methods do not perform effectively enough when meeting the transformer architecture. Because ConvNets optimize convolutional filter parameters during training, while transformer architectures learn specific pixel relationships within the training data. Compared with ConvNets, the transformer-based models are always more sensitive to the characteristics of the training data. Given these differing learning approaches, the fine tuning transfer learning strategies proposed for ConvNet-based methods are not able to be directly applied to the transformer framework. Although some recent approaches~\cite{gao2021clip,hu2022lora,sung2022lst} have been proposed to the transformer architectures, they cannot learn domain-invariant features effectively on HSIs. Because HSI datasets acquired by various sensors display distinct structural characteristics. Tuning methods designed for unimodal RGB images cannot takes advantage of the cross-modal discrepancies and handle the gaps between cross-sensor HSIs well.

In this paper, we aim to address this challenge by optimizing the external tuning strategy and internal architecture of the model. Externally, we advocate that a good representation of the target features should not only contain the knowledge from the
source domain but also preserve the target-domain-specific information. Therefore, we analyze the relationships between different sensors and propose an attention-gated tuning strategy (AGT) to reconcile conflicts in heterogeneous and cross-modal data. Specifically, we introduce a lightweight parallel auxiliary branch for the target dataset, enabling information exchange with the basic model through an attention-gated tuning block. This block uses guidance signals from the auxiliary branch and semantic information from the basic model to suppress noise and enhance relevant features, facilitating effective model tuning. Internally, for efficient feature learning from limited annotated samples, we focus on enhancing parameter utilization and flexibility in architecture design. We analyze the spatial-spectral characteristics of HSI and introduce Tri-Former, a flexible spectral-spatial separation transformer tailored for HSI classification. Tri-Former also employs 3D convolutions to enhance structural information. Our proposed method demonstrates promising performance across all three representative HSI datasets, showcasing its effectiveness in addressing the challenges of HSI classification.

The main contributions of this paper are outlined as follows.

\begin{enumerate}

    \item  A novel Attention-Gated Tuning (AGT) strategy is proposed for transformer-based HSI classification methods. To the best of our knowledge, in HSI classification field, this is the first tuning strategy that leverages transformer structures for cross-sensor and cross-modal adaptation. 
           

    \item A triplet-structured HSI classification transformer, Tri-Former, is designed after analyzing the characteristics of HSI. Tri-Former's flexible architecture enables efficient feature transfer learning from relatively limited training samples.

    \item This work establishes a connection between RGB and HSI datasets, allowing for the utilization of abundant RGB labeled datasets to enhance HSI classification performance, especially in scenarios with limited labeled HSI data.
    
\end{enumerate}

In summary, while HSIs provide rich and detailed spectral information, their complex structure necessitates advanced deep learning models for robust feature extraction. Transformer-based architectures have demonstrated outstanding performance in this regard. However, their high data requirements pose significant challenges given the limited availability of annotated HSIs. Motivated by this gap, our approach leverages transfer learning by integrating the powerful feature extraction capabilities of transformers with our novel AGT strategy and Tri-Former architecture. This integration effectively addresses the small sample problem in HSI classification. The remaining sections of this paper are structured as follows. Section II presents a review of related work. In Section III, we provide a detailed explanation of our proposed Tri-Former model and AGT strategy. Algorithm implementation details are outlined in Section IV, followed by an extensive evaluation and comparison with state-of-the-art competitors. Finally, in Section V, we conclude this work.

\section{Related Work}
\label{s2}
\subsection{ConvNets for Hyperspectral Image Classification}
ConvNets~\cite{AlexNet,vggnet,szegedy2017inception,he2016deep,huang2017densely,howard2017mobilenets,convnext} emerged as a powerful tool for visual classification, revolutionizing the field by shifting focus from handcrafted features to data-driven architecture design. The adeptness of convolutional networks in feature extraction perfectly aligns with the demands of hyperspectral image classification. The emergence of convolutional networks has also revolutionized the field of HSI classification, elevating the accuracy of classification to a new level.

Over the past decade, various convolutional variants have been proposed for HSI classification. 1D ConvNet and 2D ConvNet are first introduced into HSI classification. Mei and Ling~\cite{mei2016integrating, 2016Spectral} employed 1D ConvNets to perform convolution along the spectral dimension, effectively extracting spectral features. Meanwhile, a series of HSI classification approaches~\cite{makantasis2015deep, yue2015spectral} based on 2D ConvNet have been proposed. 1D ConvNet considers spectral dimension, while 2D ConvNet always focuses on spatial dimension. Later,  researchers pursued dual-channel ConvNet structures after recognizing the potential of combining spectral and spatial information. Zhang \textit{et al.}~\cite{zhang2017spectral} and Yang \textit{et al.} ~\cite{yang2017learning} introduced dual-channel ConvNet models that effectively integrated 1D ConvNet and 2D ConvNet, achieving further improvements in accuracy for HSI classification. Since 2016, 3D ConvNet based methods have become the mainstream as 3D convolution naturally extracts spectral-spatial features from 3D HSIs. Li \textit{et al.}~\cite{li2017spectral} and Chen \textit{et al.}~\cite{chen2016deep} pioneered this approach, constructing 3D ConvNet architectures to capture spatial-spectral patterns efficiently. The inherent 3D nature of HSIs was leveraged, enabling comprehensive feature extraction and improving classification performance. After that, researchers mainly focus on designing HSI characteristic-aligned 3D ConvNets. Efficient residual structures were introduced by Zhong \textit{et al.}~\cite{zhong2018spectral}, integrating spectral and spatial residual modules to improve feature learning. Zhang \textit{et al.}~\cite{2019Hyperspectral} focused on lightweight 3D-CNN designs and transfer learning strategies to handle small sample problems. Getting inspiration from DARTS, Chen \textit{et al.}~\cite{chen2019automatic} proposed a 3D Auto-CNN for HSI classification. In the preprocessing stage, 3D Auto-CNN heavily compresses the spectral dimension of raw HSIs through point wise convolution. Under this basis, 3DANAS~\cite{zhang20213} and HyT-NAS~\cite{3danas} are designed, which introduce the NAS algorithm into an efficient 3D pixel-to-pixel classification framework.

The evolution of ConvNet-based HSI classification methods embodies three fundamental principles, including: 1) simultaneously harnessing spectral and spatial information contributes to enhancing accuracy; 2) the 3D convolution operation is inherently well-suited for processing 3D HSI; 3) aligning the classification framework with the inherent characteristics of HSI is unignorable. These three guiding principles have been adhered to in the conception of the Tri-Former model.

\subsection{Vision Transformers for Hyperspectral Image Classification}

Transformer model was initially introduced for Natural Language Processing (NLP) by Vaswani et al.~\cite{vaswani2017attention}. It brought about a revolutionary change in processing sequential data through its attention mechanism. In the field of computer vision, ViT~\cite{dosovitskiy2020image} builds on the success of the transformer, which was originally designed for NLP.

More recently, within the HSI classification domain, researchers have redirected their focus towards the transformer model in their quest for enhanced performance~\cite{ hong2021spectralformer,he2021spatial, qing2021improved}. For instance, He \textit{et al.}~\cite{he2021spatial} introduced a spatial-spectral transformer that combines a ConvNet for capturing spatial information with a ViT for extracting spectral relationships. Similarly, the method proposed in~\cite{qing2021improved} adopts a spectral relationship extraction transformer along with several decoders. SpectralFormer~\cite{hong2021spectralformer} mines and represents the sequence attributes of spectral signatures well from input HSIs using the transformer model. It extracts spectrally local sequence information from neighboring bands of HSIs. Meanwhile, HyT-NAS~\cite{3danas} first combines the NAS and transformer for handling the HSI classification task. An emerging transformer module is grafted on the automatically designed ConvNet to add global information to local region focused features. The searching space in HyT-NAS is hybrid, suitable for the HSI data with a relatively low spatial resolution and an extremely high spectral resolution. To better utilize  high-level semantic spectral-spatial features, a spectral-spatial feature tokenization transformer (SSFTT) method~\cite{sun2022spectral} is proposed. Specifically, a spectral–spatial feature extraction module is built to extract low-level shallow features and a gaussian weighted feature tokenizer is introduced for feature transformation. Based on this, the transformed features are input into the transformer encoder module for stronger feature representation and learning. Another model GraphGST~\cite{gst} incorporates a new absolute positional encoding (APE) to capture pixel positional sequences and uses self-supervised learning for cross-view contrastive learning, improving representation and capturing local-to-global correlations in HSI classification tasks.

The methods mentioned above have shown the transformer's potential in effectively leveraging spatial and spectral relationships within hyperspectral data. However, there is still significant room for improvement. In this paper, we further enhance the intrinsic strengths of the transformer architecture in HSI classification by improving its parameter utilization and flexibility.

\subsection{Fine Tuning Methods}

In real-world scenarios, obtaining a sufficient amount of training data for HSI classification is challenging due to the expensive sensor cost and labeling cost. Therefore, designing models with low training sample requirements has consistently been a goal within the HSI classification domain. A classical approach to addressing this challenge is to pretrain the method on the sufficiently labeled data (maybe from another domain) first, and then transfer the  acquired  knowledge and representations to the target domain~\cite{pan2009survey}. 

\noindent\textbf{Fine Tuning ConvNets in HSI classification.} As a pioneering effort to address HSI classification problems with limited training samples, Yang \textit{et al.} ~\cite{ yang2017learning} introduced the transfer learning strategy into a two-branch ConvNet to extract more relevant features from HSIs. In~\cite{2019Hyperspectral}, Zhang \textit{et al.} further improved this idea and transferred feature extraction capacity among heterogeneous HSI datsets and even cross-modal datasets. Unfortunately, these methods may not exhibit effective performance on transformer architecture. In LWNet~\cite{2019Hyperspectral} and Two-CNN~\cite{yang2017learning}, fine tuning strategies are employed on ConvNet architectures, yielding promising results. However, unlike ConvNets, which learn convolutional filter parameters from extensive datasets, transformer-based methods learn specific pixel relationships within the training data. Given these differing learning approaches, the fine tuning transfer learning strategy used in ConvNet-based methods exhibit limited effectiveness in the transformer framework. 

\begin{figure}
	\begin{center}
		\includegraphics[width=0.95\linewidth]{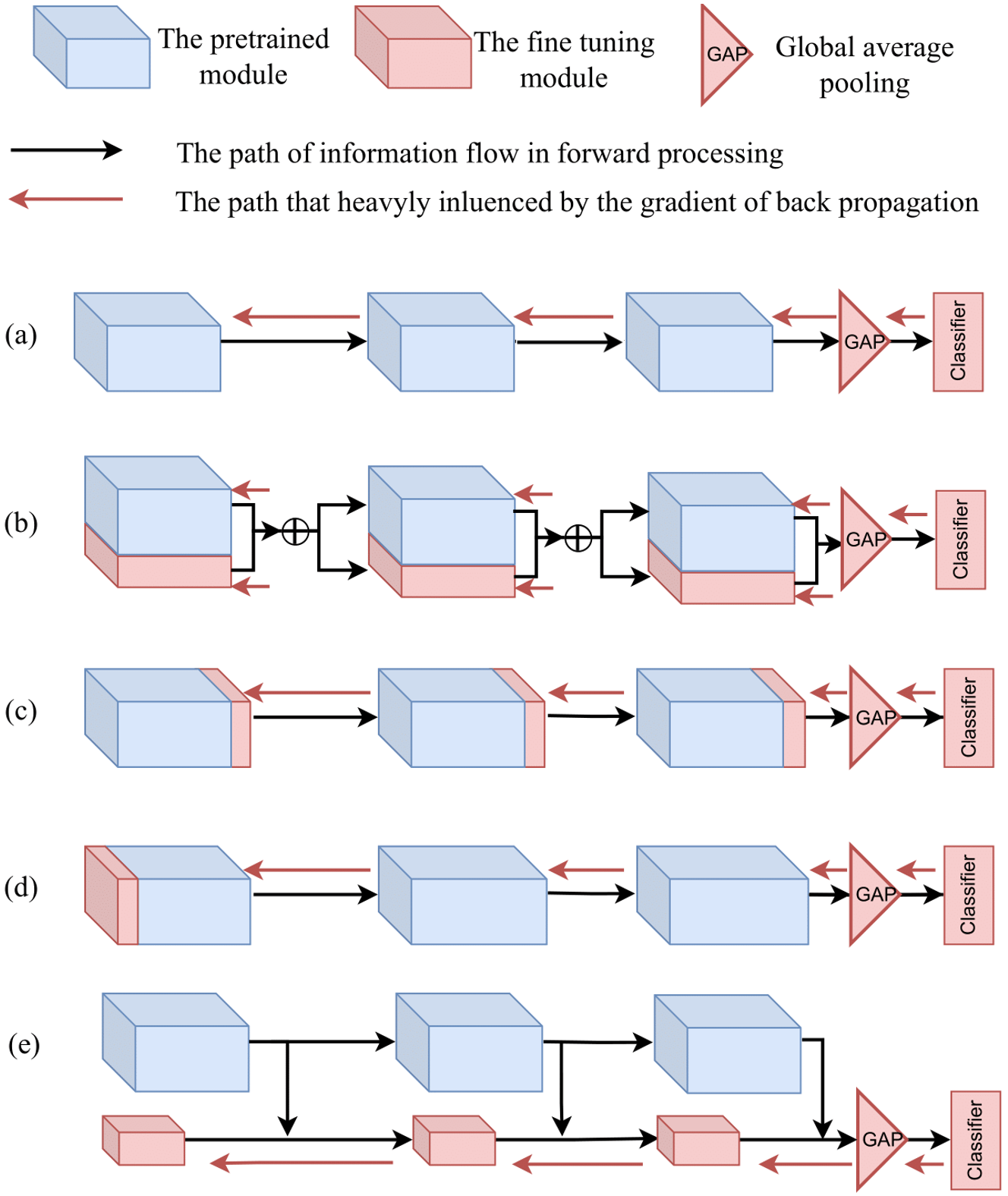}
        \end{center}
    \vspace{-0.3cm}
	\caption{Comparison of different fine tuning architectures. (a) FPFT; (b) Adapter tuning; (c) LoRA; (d) Prompt tuning;  (e) Ladder side tuning. }
	\label{fig:related work}
 \vspace{-0.3cm}
\end{figure}

\noindent\textbf{Full-Parameter Fine Tuning.} Due to the rapid advancements in vision language models, an increasing number of fine tuning methods targeting transformer architectures have been proposed. As shown in Fig.~\ref{fig:related work}(a), Full Parameter Fine Tuning (FPFT) adjusts all layers and parameters based on the pre-trained model to adapt it to a specific task. This process usually uses a small learning rate and task-specific data to fully utilize the general features of the pre-trained model, but may require more computing resources. 

\noindent\textbf{Parameter-Efficient Fine Tuning.} Parameter-Efficient Fine Tuning (PEFT) technology aims to improve the performance of pre-trained models on new tasks by minimizing the number of fine tuning parameters and computational complexity. In this way, even if computing resources are limited, the knowledge of the pre-trained model can be used to quickly adapt to new tasks and achieve efficient transfer learning. Therefore, PEFT technology can greatly shorten the model training time and computational cost while improving the model effect, allowing more people to participate in deep learning research. A popular line of work on PEFT is to add a few trainable parameters without changing the original parameters of the pre-trained model. Fig.~\ref{fig:related work}(b)-Fig.~\ref{fig:related work}(e) illustrate the information forward and gradient backpropagation process of some representative PEFT approaches. Adapter tuning~\cite{gao2021clip} approaches always insert small neural network modules, called "adapters", between each layer of the model or between certain specific layers. These adapters are trainable, while the parameters of the original model remain unchanged. Similar to Adapter tuning, LoRA~\cite{hu2022lora} adds small, low-rank matrices to the key layers of the model to adjust the model's behavior. Prompt tuning~\cite{Prompttuning} adds learnable embedding vectors as prompts to the input of the pre-trained language model. These prompts are designed to be updated during training to guide the model to output more useful responses for specific tasks. Without inserting additional parameters inside backbone networks, another line of work on PEFT is to add  an additional small and separate network. Zhang~\textit{et al.}~\cite{zhang2020side} proposed Side-tuning, which adapts a pre-trained network by training a lightweight “side” network that is fused with the (unchanged) pre-trained network via summation. In a similar trend, LST~\cite{sung2022lst} trains a ladder side network. This small and separate network takes intermediate activations as input via shortcut connections (called ladders) from backbone networks and makes predictions.

In this paper, in order to make full use of labeled cross-sensor data, we propose an attention-gated tuning (AGT) strategy and employ it to fintune the proposed efficient Tri-Former. The proposed approach demonstrates effectiveness across homogeneous, heterogeneous, and even cross-modality cases. Although the proposed AGT takes inspiration and has a similar parallel structure to LST, we argue that there are major differences in motivations, key problems and architecture designs between the two methods. Detail comparisons and  comprehensive analysis are presented in Section IV.C. The corresponding experimental results are outlined in Section IV.J-IV.I.

\section{Proposed Method}
In this section, we provide a detailed introduction to our Tri-Former model and the proposed Attention-Gated Tuning (AGT) strategy. 
We first briefly revisit the vanilla Vision Transformer (ViT) architecture, then describe the Tri-Former architecture, introducing  how the Spectral-Spatial Parallel (SSP) block and 3D convolution reduce redundancy and enhance learning for HSI data. Following that, we elaborate on the details of the AGT strategy, emphasizing its ability to leverage cross-sensor or cross-modality data and differences from previous approaches. 
Finally, following~\cite{feng2024class,ding2022self,ding2022unsupervised,zhang2023multireceptive}, we provide the pseudo code of the working flow of proposed method in Algorithm \ref{alg:tri_former_agt_en}.

\subsection{Primary introduction of ViT model}
ViT is a representative vanilla visual transformer model. It directly processes images as sequences of patches, and then utilizes self-attention mechanisms to capture long-range dependencies between these patches. The self-attention operation computes a weighted sum of the embeddings based on their pairwise relationships, which can be expressed as follows:
\begin{equation}
{Attention(Q,K,V)}=\operatorname{softmax}\left(\frac{Q K^{\top}}{\sqrt{d_{k}}}+P\right) V
\end{equation}
where $Q$,$K$,$V$ are the query, key, and value matrices, respectively, representing the input embeddings. $d_k$ denotes the dimensionality of feature, and $P$ denotes the relative position embedding term. 

Based on self-attention operation, the multi-head attention layer can be constructed. Each transformer block consists of two sub-layers: a Multi-Head Self-Attention (MHSA) layer (token mixer) and a Feed-Forward Neural Network (FFN) layer (channel mixer). The output of each transformer block is then fed to the subsequent block, enabling hierarchical feature extraction.

Apart from the aforementioned layer, the ViT model also includes a special learnable positional embedding that encodes the position information of each patch in the image sequence. Two common methods of adding position embedding are: 1) applying RPB to the attention map, as in Eq.1; 2) adding absolute position embedding to the patch embeddings. In the vanilla ViT model, preserving positional information is crucial. However, employing position embeddings can be a bit cumbersome, especially when there are changes in the input patch resolution. As shown in Eq.1, in this paper, we adopt the former one. The RPB $P_{(x,y),(x',y')}^{m}$ is:


\begin{equation}
\label{eq:rpb}
P_{(x,y),(x',y')}^{m}
= B_{|x-x'|,|y-y'|}^{m},
\end{equation}
where $B^{m}\in\mathbb{R}^{(2h-1)\times(2w-1)}$ represents the translation-invariant attention bias for the head $m$, with $h,w$ being the spatial dimensions of the feature map.

\begin{figure}
	\begin{center}
		\includegraphics[width=1.0\linewidth]{ 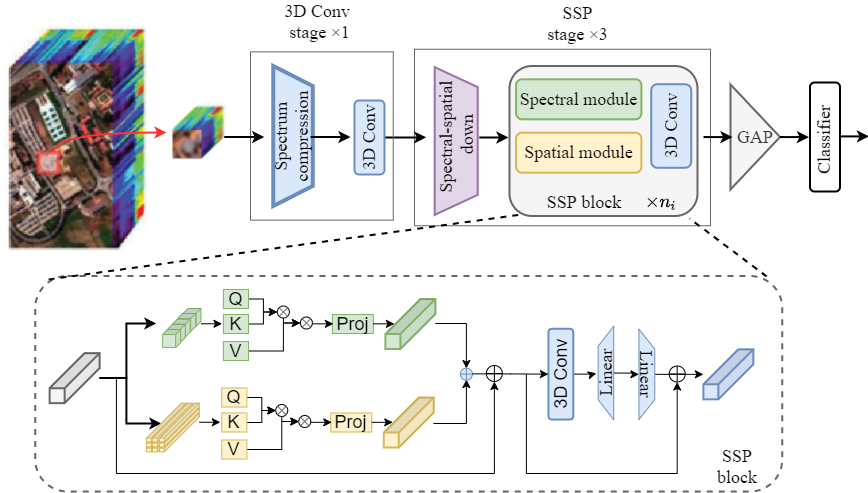}
	\end{center}
	\caption{Architecture of proposed Tri-Former. SSP block has three main parts: spectral module, spatial module and 3D convolution module. Spectral module is responsible for collecting information from different spectrum bands. Spatial module is in charge of collecting information from different space locations. 3D convolution layer is added before the two linear layers to enforce 3D structure information and stabilize training process. }
	\label{fig:triformer}
\end{figure}


\subsection{Architecture of Tri-Former}
To efficiently learn features from limited annotated samples, Tri-Former emphasizes improving architectural flexibility and reducing redundant computations. By analyzing the spatial-spectral characteristics of HSIs, we propose Tri-Former, a triplet-structured transformer designed specifically for HSI classification. Tri-Former employs the spectral-spatial parallel architecture to tackle computational inefficiencies and incorporates 3D convolution to enhance structural information.

\noindent\textbf{Overall architecture.} As is depicted in Fig. \ref{fig:triformer}, following the common setting adopted by ResNet ConvNext, etc. the overall framework consists of four stages. Inspired by FcaFormer \cite{zhang2023fcaformer}, LeViT \cite{graham2021levit}, etc, we adopt early convolution architecture. The overall architecture is one 3D convolution stage followed by three transformer stages, as shown in the top half of Fig. \ref{fig:triformer}. Stage one comprises a spectrum compression module and a 3D convolution module. The subsequent three stages are constructed with the proposed spectral-spatial parallel (SSP) blocks. Each stage begins with a spectral-spatial downsample layer, followed by $n_{i}$ SSP blocks. Following the classical design strategy, the widths of the four stages are progressively doubled from stage to stage. In this work, the width of the first stage is set to 32 as the starting point for subsequent widening. This design choice enables the network to efficiently process and extract relevant information in a computationally efficient manner. 

\noindent\textbf{Spectral-spatial parallel block.} SSP block first compresses the spectral dimension of input feature. After that, each feature map $f\in {\mathbb{R}}^{c\times s \times w \times h}$ is splitted to sequence and input to the spatial module and spectral module.  

Specifically, for input feature $f\in\mathbb{R}^{c\times s\times w\times h}$ where $c$=channels, $s$=spectral bands, $(w,h)$=spatial dimensions:

\begin{itemize}
\item \textbf{Spectral attention} splits features along spectral dimension:
\begin{equation}
\begin{aligned}
Q_e &= f_W^Q(f) \in \mathbb{R}^{R_e \times N_e \times L_e} \\
K_e &= f_W^K(f) \in \mathbb{R}^{R_e \times N_e \times L_e} \\
V_e &= f_W^V(f) \in \mathbb{R}^{R_e \times N_e \times L_e}
\end{aligned}
\end{equation}
where $R_e=w\times h$ (flattened spatial positions), $N_e=s$ (spectral bands), $L_e=c$ (channel dimension).

\item \textbf{Spatial attention} splits features along spatial dimensions:
\begin{equation}
\begin{aligned}
Q_a &= f_W^Q(f) \in \mathbb{R}^{R_a \times N_a \times L_a} \\
K_a &= f_W^K(f) \in \mathbb{R}^{R_a \times N_a \times L_a} \\
V_a &= f_W^V(f) \in \mathbb{R}^{R_a \times N_a \times L_a}
\end{aligned}
\end{equation}
where $R_a=s$ (spectral bands), $N_a=w\times h$ (spatial positions), $L_a=c$ (channel dimension).
\end{itemize}

The attention computation follows:
\begin{equation}
\begin{aligned}
\mathrm{Atten}_{spe} &= \mathrm{LayerNorm}\left(\mathrm{softmax}\left(\frac{Q_eK_e^{\top}}{\sqrt{d_e}}\right)V_e\right) \\
\mathrm{Atten}_{spa} &= \mathrm{LayerNorm}\left(\mathrm{softmax}\left(\frac{Q_aK_a^{\top}}{\sqrt{d_a}}\right)V_a\right)
\end{aligned}
\end{equation}

where $\mathrm{softmax}$ applies along the second dimension $N$, and $d$ is the scaling factor corresponding to $L$ dimension.

After these attention computations, we reshape $\mathrm{Atten}_{\mathrm{spe}}$ and $\mathrm{Atten}_{\mathrm{spa}}$ back to $\mathbb{R}^{c \times s \times w \times h}$ and combine them with a residual connection:
\begin{equation}
\label{eq:f_ssp}
f_{\mathrm{ssp}}
= f
+ \mathrm{reshape}\bigl(\mathrm{Atten}_{\mathrm{spe}})
+ \mathrm{reshape}\bigl(\mathrm{Atten}_{\mathrm{spa}})
\end{equation}
To further enhance structural consistency and stabilize training, we apply a 3D convolution followed by an MLP:
\begin{equation}
\label{eq:f_out}
f_{\mathrm{out}}
= f_{\mathrm{ssp}}
+ \mathrm{MLP}\Bigl(\mathrm{Conv3D}\!\bigl(f_{\mathrm{ssp}}\bigr)\Bigr).
\end{equation}
Here, $\mathrm{Conv3D}$ denotes a 3D convolution (e.g., kernel size $3\times3\times3$). 
This design effectively embeds position-like information without relying on explicit positional embeddings, reducing the overhead when input patch size changes.

In Eq.7, a 3D convolution is employed. Such a design has two advantages: 1) this additional 3D convolution contributes to stabilize the optimization process of the transformer block; 2) it enhances the structure information (position information) and obviates the necessity to rely on position embedding in the original ViT.

After the sequential 3D Conv and SSP stages, the learned features are transformed into feature vectors using global average pooling (GAP). Finally, a classifier is used to generate the final prediction.
\begin{figure}
	\begin{center}
		\includegraphics[width=0.8\linewidth]{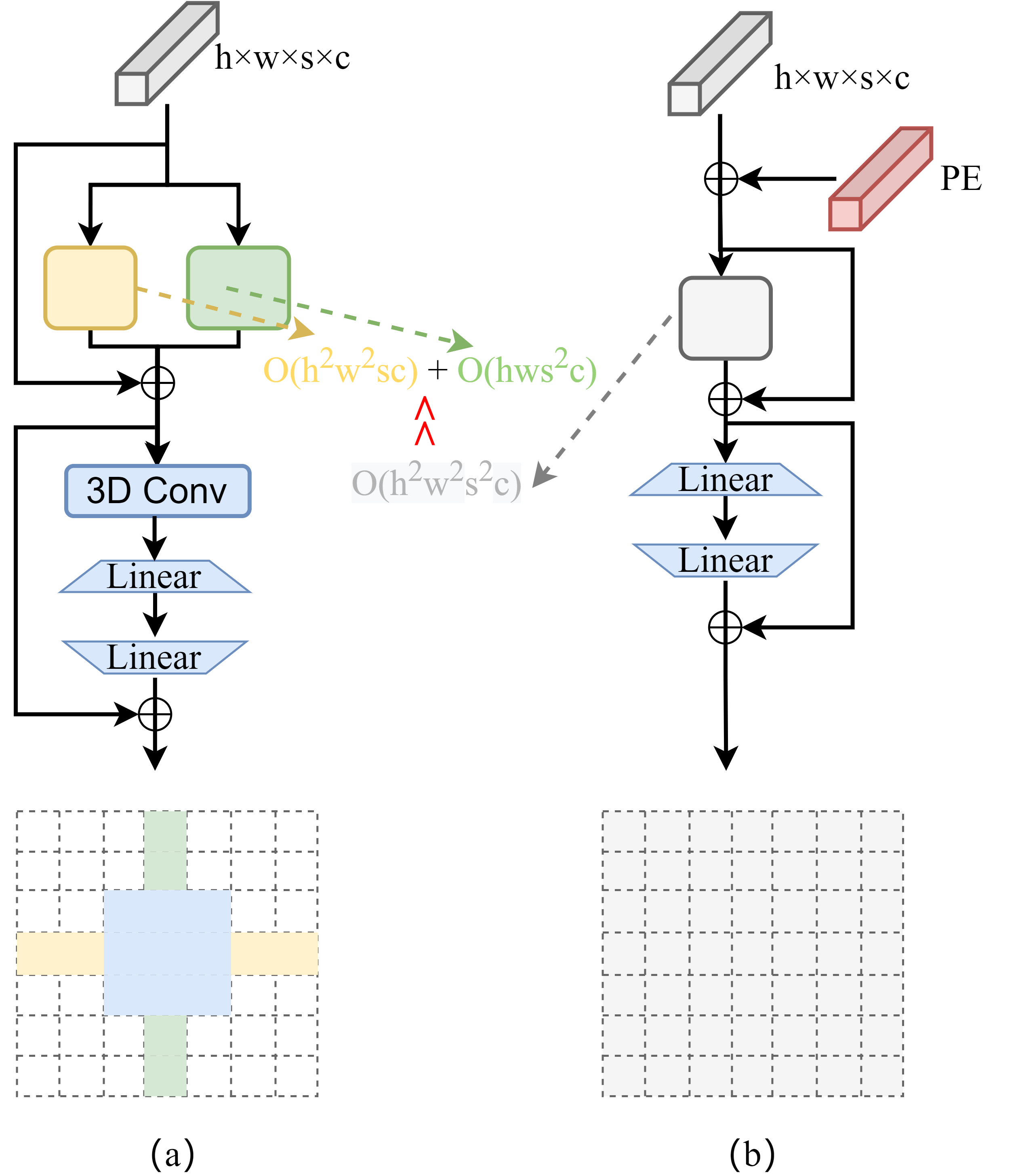}
	\end{center}
	\vspace{-0.3cm}
	\caption{Comparison between proposed SSP block and vanilla ViT block. (a) SSP block; (b) Vanilla ViT block.}
	\label{fig:Triformer_vs_vit}
\end{figure}

\begin{figure*}
	\begin{center}
		\includegraphics[width=0.9 \linewidth]{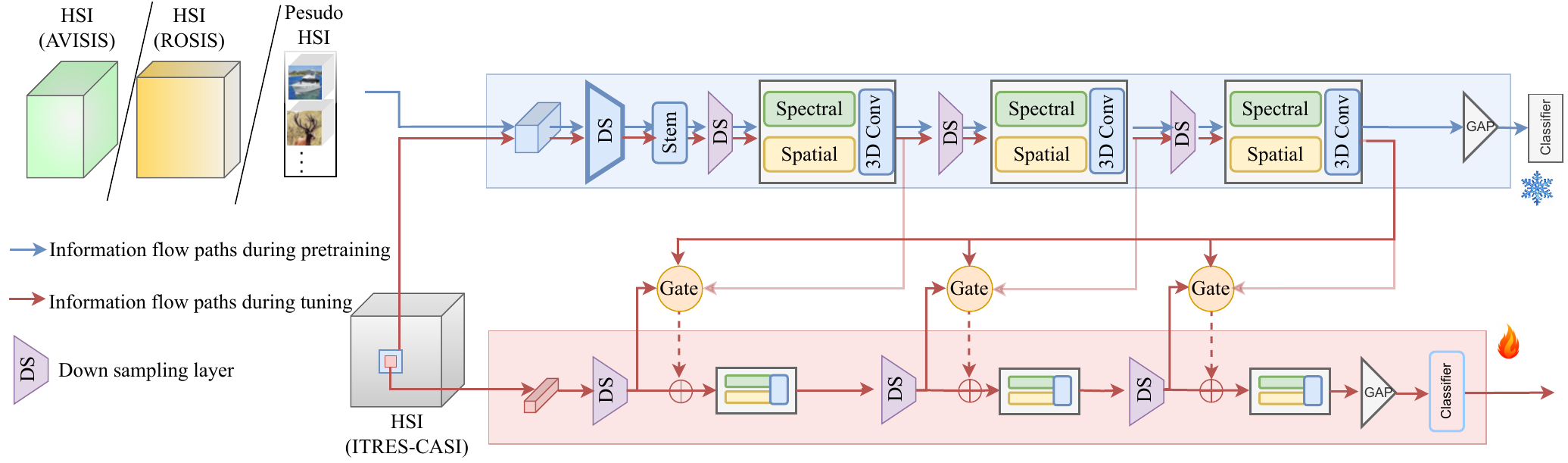}
	\end{center}
	\vspace{-0.3cm}
	\caption{Architecture of proposed AGT. Blue part is the basic model, where big patches and heavy Tri-Former are adopted. Red part is an auxiliary branch, where a small patch and tiny Tri-Former are used.}
	\label{fig:sdt}
 \vspace{-0.3cm}
\end{figure*}

\noindent\textbf{Analysis on computational complexity.} The proposed spectral-spatial parallel (SSP) block significantly reduces the computation cost. Instead of directly applying self-attention on three dimensions (two spatial dimensions and one spectral dimension), spectral and spatial information are separately processed. The computation complexity of the vanilla ViT is quantic complex regarding to the length of the input sequence. In 3D architecture HSI, this issue is more serious. As shown in Fig.~\ref{fig:Triformer_vs_vit}, for a input feature $f\in {\mathbb{R}}^{c\times s \times w \times h}$, if we directly compute self-attention on three dimensions, the computation complexity is:
\begin{equation}
O(\operatorname{VANILLA}(f))=O(h^{2}w^{2}s^{2}c)
\end{equation}
where VANILLA denotes the vanilla transformer architecture that directly applies self-attention on three dimensions. In the proposed SSP architecture, the computational complexity becomes:
\begin{equation}
O(\operatorname{SSP}(f))=O(h^{2}w^{2}sc)+O(hws^{2}c)
\end{equation}
In practical experiments, the value of $c$ and $s$ are relatively large, making $O(\operatorname{SSP}(f))$ significantly smaller than $O(\operatorname{VANILLA}(f))$. For instance, if the input feature map has a spatial size of $27 \times 27$, with 32 channels and 144 spectral bands. $O(\operatorname{SSP}(f))=O(2.93\times 10^{9})$, while $O(\operatorname{VANILLA}(f))=O(3.53 \times 10^{11})$. 

Obviously, the designed SSP block significantly reduces  computational load, rendering the entire architecture more lightweight and adaptable, This beneficial to feature learning on HSIs with limited  annotations.

\noindent\textbf{Analysis on receptive field.}  The enhancements on transformer architecture have significant impacts on their receptive field. The receptive field in the proposed SSP block excludes certain irrelevant elements within its scope, consequently strengthening the attention on pixels with close relationships. Using two-dimensional features as an example, we illustrate the receptive fields of SSP block and that of vanilla transformer in Fig.~\ref{fig:Triformer_vs_vit}. From Fig.~\ref{fig:Triformer_vs_vit}(b), it is evident that the receptive field of the vanilla structure is a square region. But as in Fig.~\ref{fig:Triformer_vs_vit}(a), the proposed SSP block considers pixels that share the same row, column, or spatially adjacent positions with the central pixel. The self-attention operation in the ViT architecture achieves information fusion by computing approximate similarities between each pair of pixels. However, when a pixel is relatively distant from the central pixel, it may not effectively contribute to extracting information from the central pixel. Thus, retaining such distant pixels within the receptive field may indeed introduce noise. Therefore, the receptive field of our proposed module retains only the most crucially related pixels, while excluding irrelevant interference pixels, resulting in a more efficient mechanism.

\subsection{Attention-Gated Tuning }
Attention-Gated Tuning (AGT) is proposed to enable a transformer-based HSI classification method to leverage cross-sensor or cross-modality data to further enhance performance. As illustrated in Fig.~\ref{fig:sdt}, AGT serves as a bridge, bridging the gaps between target HSIs and images captured by other sensors. 
The process of AGT contains two sequential stages:
\begin{itemize}
    \item Firstly, the basic model is pretrained  with homogeneous and heterogeneous hyperspectral data, even RGB data, as illustrated in the blue part of Fig.~\ref{fig:sdt}. It is worth noting that there is a significant modal disparity between RGB data and hyperspectral data. Therefore, we initially generate pseudo-hyperspectral data from RGB data to mitigate these modality gaps. Specifically, we employ  DRCR-Net\footnote{\url{https://github.com/jojolee6513/DRCR-net}} to create the 32-channel pseudo-HSI data from RGB data~\cite{li2022drcr} and then use these pseudo-HSIs to perform pretraining. 
    These generated images possess certain spectral characteristics while harnessing the advantages of extensive natural image data. Experimental results in Section~IV.I demonstrate that the AGT can also benefit from RGB data.

    \item Secondly, a tiny auxiliary branch is introduced to the basic model before tuning on the target HSI datasets, as presented in the Fig.~\ref{fig:sdt}. Corresponding to the $n_i$ blocks in the basic model, the auxiliary branch consists of $n_i$ lightweight SSP blocks. To reduce the overhead associated with the auxiliary branch, we choose smaller patches, shallower architectures, and narrower modules. Through an attention-gated strategy, the proposed AGT can autonomously select useful information from the basic model to guide auxiliary branch learning. Unlike previous fine tuning methods~\cite{jia2022visual, liu2023pre, liu2021p,sung2022lst}, where the basic model remains frozen during tuning, we employ an asynchronous cold-hot gradient update strategy, in which the basic model is updated slowly. 
\end{itemize}

\begin{figure}[t]
	\begin{center}
		\includegraphics[width=0.7 \linewidth]{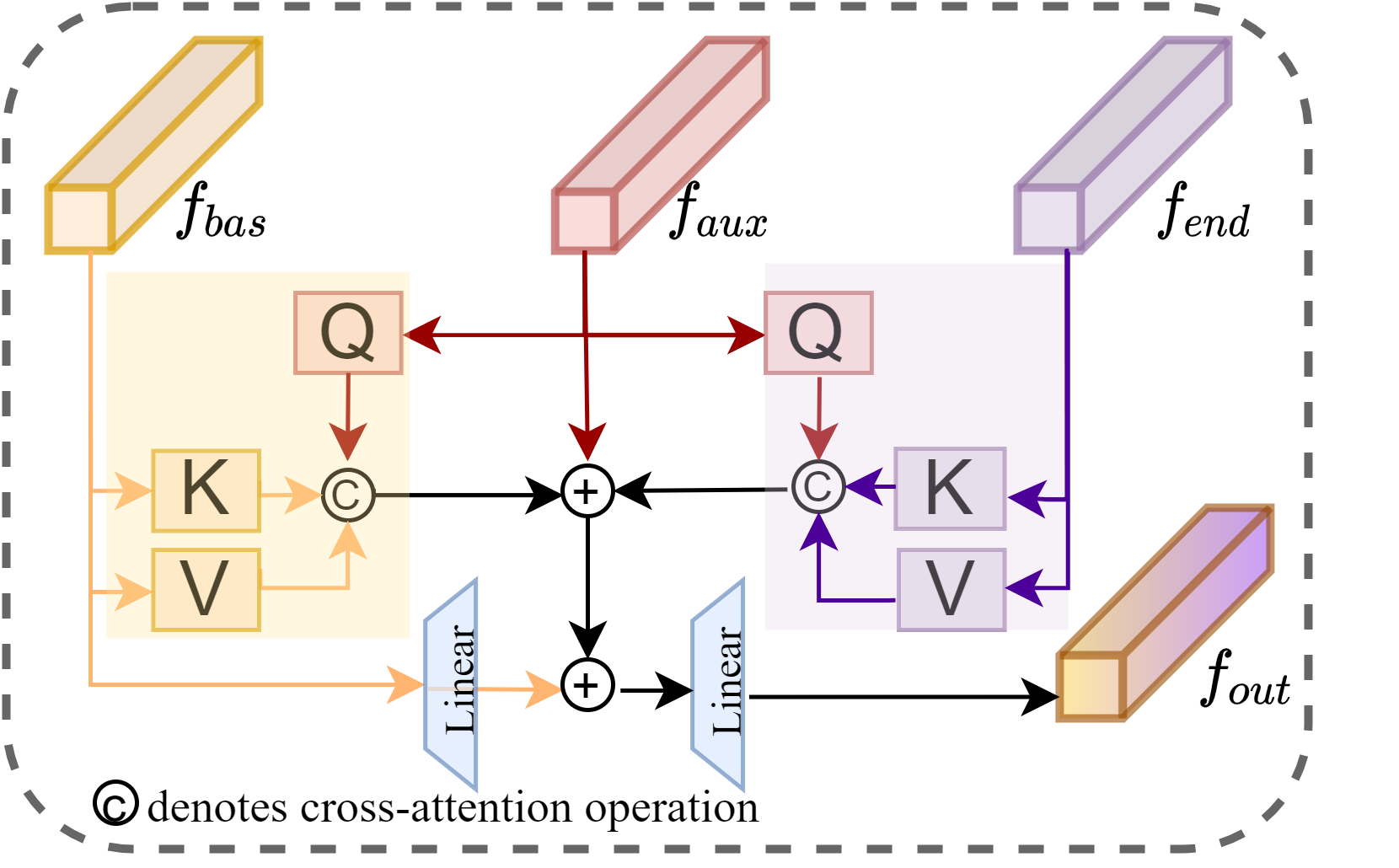}
	\end{center}
	\vspace{-0.3cm}
	\caption{Internal structure of proposed attention gate.}
    \vspace{-0.5cm}
	\label{fig:agt}
\end{figure}

Fig.~\ref{fig:agt} shows the internal structure of the attention gate. The auxiliary branch feature \( f_{\text{aux}} \) is used to generate the query, while the basic model feature \( f_{\text{bas}} \) and the final stage feature \( f_{\text{end}} \) construct the key and value:

\begin{equation}
Q_{\text{aux}} = W_q \cdot f_{\text{aux}},
\end{equation}

\begin{equation}
K_{\text{bas}} = W_k \cdot f_{\text{bas}}, \quad V_{\text{bas}} = W_v \cdot f_{\text{bas}},
\end{equation}

\begin{equation}
K_{\text{end}} = W_k \cdot f_{\text{end}}, \quad V_{\text{end}} = W_v \cdot f_{\text{end}},
\end{equation}

where \( W_q, W_k, W_v \) are learnable projection matrices. The auxiliary branch selectively integrates information from both the basic model and the final stage using two separate cross-attention modules:

\begin{equation}
A_{\text{bas}} = \operatorname{softmax} \left( \frac{Q_{\text{aux}} K_{\text{bas}}^\top}{\sqrt{d}} \right) V_{\text{bas}},
\end{equation}

\begin{equation}
A_{\text{end}} = \operatorname{softmax} \left( \frac{Q_{\text{aux}} K_{\text{end}}^\top}{\sqrt{d}} \right) V_{\text{end}}.
\end{equation}

The final representation is computed as:

\begin{equation}
A_{\text{gate}} = A_{\text{bas}} + A_{\text{end}}+f_{\text{aux}},
\end{equation}

\begin{equation}
f_{\text{out}} = W_{\text{linear}} \cdot (W_{\text{linear}} \cdot f_{\text{bas}} + A_{\text{gate}}).
\end{equation}
where $W_{\text{linear}}$ is learnable Linear projection matrices.The gating mechanism effectively selects relevant information from both shallow and deep feature representations, ensuring robust knowledge transfer across sensors and modalities.

\begin{algorithm}[t]
\caption{\textbf{Pseudo-code of Tri-Former with AGT}}
\label{alg:tri_former_agt_en}

\textbf{Input:}\\
    1)~Source dataset $\mathcal{D}_S = \{(x_i^{(S)}, y_i^{(S)})\}$ \\
    2)~Target dataset $\mathcal{D}_T = \{(x_j^{(T)}, y_j^{(T)})\}$ \\
    3)~Network hyper-parameters (e.g., patch size, channels $c$, spectral bands $s$, etc.)\\
\textbf{Output:}Trained model parameters $(\theta_{\mathrm{base}},\theta_{\mathrm{aux}})$
\vspace{4pt}
\hrule 
\vspace{4pt}
\textbf{Stage A: Pretrain the Base Tri-Former Model} \\
\textbf{Initialize} base model parameters $\theta_{\mathrm{base}}^{(0)}$\;

{Pre-training on {mini-batch $\{(x_i^{(S)}, y_i^{(S)})\}\!\subset\!\mathcal{D}_S$}{\\
    1)~\textbf{Forward pass} (cf. Eq.(1)-(7)) \\
    2)~\textbf{Compute loss:}
    \(\displaystyle
      \mathcal{L}_{S} = \mathrm{CE}\!\bigl(\hat{y}_i^{(S)},\, y_i^{(S)}\bigr)
    \)\;

    3)~\textbf{Backward and update} base model:
    \(\displaystyle
      \theta_{\mathrm{base}}
      \leftarrow
      \theta_{\mathrm{base}}
      - \eta\,\nabla_{\theta_{\mathrm{base}}}\,\mathcal{L}_{S}
    \)
}

\vspace{4pt}

\hrule 
\vspace{4pt}
\textbf{Stage B: Integrate Auxiliary Branch and Perform AGT} \\
\textbf{Initialize} auxiliary branch parameters $\theta_{\mathrm{aux}}^{(0)}$\ 

Tuning on target dataset {mini-batch $\{(x_j^{(T)}, y_j^{(T)})\}\!\subset\!\mathcal{D}_T$}{\\
    1)~\textbf{Forward pass} (cf. Eq.(10)--(16)):

    2)~\textbf{Compute target loss}:
    \(\displaystyle
      \mathcal{L}_{T}
      =
      \mathrm{CE}\bigl(\hat{y}_j^{(T)},\, y_j^{(T)}) 
    \)\;

    3)~\textbf{Asynchronous cold-hot updates}:
    \begin{align*}
      &\theta_{\mathrm{aux}}
      \;\leftarrow\;
      \theta_{\mathrm{aux}}
      \;-\;
      \alpha\,\nabla_{\theta_{\mathrm{aux}}}\!\mathcal{L}_{T}
      \quad(\text{hot: larger LR } \alpha)\\
      &\theta_{\mathrm{base}}
      \;\leftarrow\;
      \theta_{\mathrm{base}}
      \;-\;
      \beta\,\nabla_{\theta_{\mathrm{base}}}\!\mathcal{L}_{T}
      \quad(\text{cold: smaller LR } \beta \ll \alpha)
    \end{align*}
}

\vspace{4pt}
 \textbf{Return:} $(\theta_{\mathrm{base}}, \theta_{\mathrm{aux}})$
 }
\end{algorithm}


In terms of design, AGT has three notable aspects that guarantee its efficiency in our scenario. To elaborate further:

\begin{itemize}
\item \textbf{Parallel architecture}. In the proposed AGT approach, a tiny auxiliary branch is introduced and updated to adapt to the target datasets. This modification arises from the acknowledgment that distinct HSI datasets exhibit more pronounced variations when compared to the disparities between diverse RGB datasets. Incorporating only a limited number of new learnable parameters within the basic model has proven insufficient for effectively bridging this gap. Such a parallel architecture offers two distinct advantages. Firstly, during the tuning process, detaching the foundational model from gradient updating becomes effortless, enabling a more controlled adjustment. This adjustment aligns perfectly with the requirements for fine tuning the transformer model for HSI classification. Secondly, the architectural design of the auxiliary branch is more flexible.

\item \textbf{Attention-guided learning.} In the proposed AGT, features from each stage of the basic model are introduced to guide the auxiliary branch learning. Specifically, as shown in Fig.~\ref{fig:agt}, $f_{bas}$ and $f_{end}$ denote the feature from current stage and last stage of basic model, respectively. Shallow features contain an excess of detailed information, whereas deep features encompass more semantic information. AGT combines shallow and deep features learned from pre-training to provide stronger guidance for HSI classification tasks. The gating mechanism is achieved through cross-attention, as shown in Fig.~\ref{fig:agt}. This approach allow the auxiliary branch to selectively fuse useful information from the basic model while filtering out conflict information learned during pretraining.

\item \textbf{Asynchronous cold-hot gradient update strategy}. From comprehensive experiments, we find that completely freezing the foundational model as what is done in RGB image processing and NLP fields~\cite{jia2022visual, liu2023pre} does not work well in HSI classification. We speculate that this phenomenon arises due to the significant disparities among different HSI datasets. Therefore, a more suitable approach involves gradually fine tuning the basic model, allowing it to adjust appropriately and accommodate new data. Therefore, we propose an asynchronous cold-hot gradient update strategy, where \textcolor{blue}{basic model} is updated slowly (\textcolor{blue}{cold)} and the \textcolor{red}{auxiliary branch} is updated quickly (\textcolor{red}{hot}).

\end{itemize}

\begin{figure}[t]
	\begin{center}
		\includegraphics[width=0.8 \linewidth]{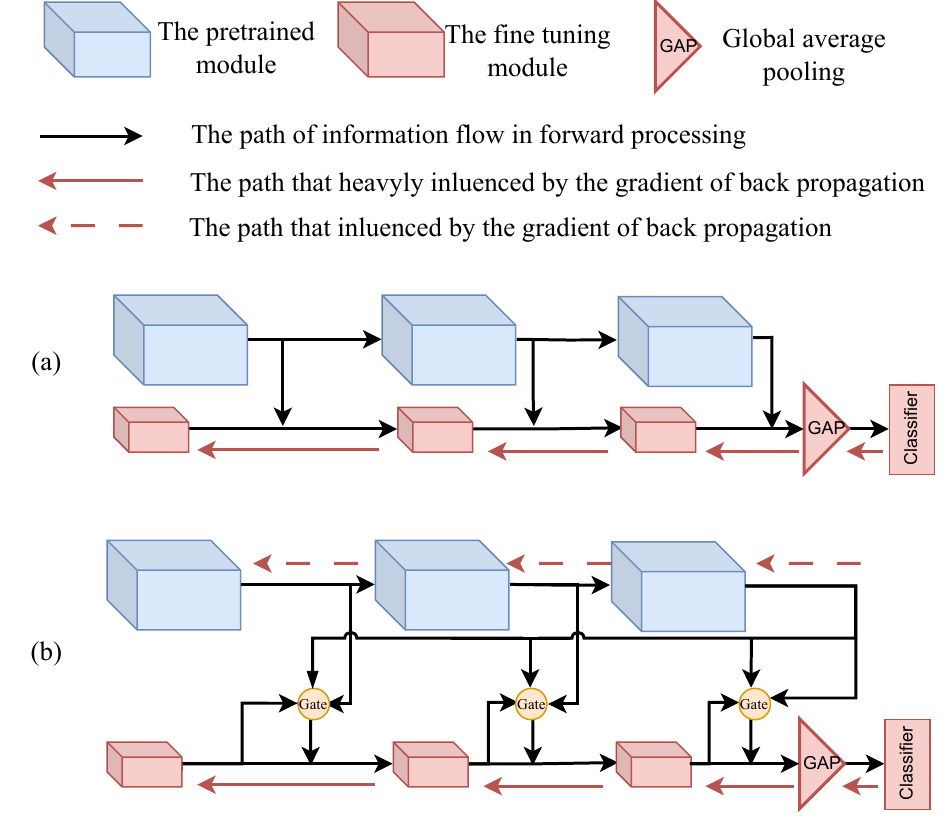}
	\end{center}
	\vspace{-0.3cm}
	\caption{Comparison on proposed AGT and related work. (a). LST; (b). AGT.}
 \vspace{-0.5cm}
	\label{fig:TF_vs}
\end{figure}

Although the design of AGT is inspired by the core idea of ladder side tuning (LST)~\cite{jia2022visual}, their motivations, key problems, architectures and gradient updating are totally different. 1) AGT aims to alleviate limited label sample problems in HSI classification, whereas LTS focuses on fine tuning large language models (LLM); 2) AGT deals with HSIs collected by various sensors or even datasets from different domains. LST is proposed for datasets from the same modality; 3) AGT uses information from the auxiliary branch to guide feature fusion, which reuse the features from the current stage and the end stage; 4) As depicted in Fig.~\ref{fig:TF_vs}(b), AGT employ a cold-hot gradient update strategy , while LST always freeze the basic model. The corresponding experiments in Section IV.G demonstrate the effectiveness of proposed AGT. 

\subsection{Theoretical Analysis from Domain Adaptation Perspective}

In this section, we provide a theoretical perspective on \textit{Attention-Gated Tuning (AGT)} from the perspective of \textit{domain adaptation}. Cross-sensor or cross-modality hyperspectral data often exhibit significant discrepancies from the data used to pre-train the backbone model. Consequently, naive fine-tuning (either fully or partially) on the new target dataset may lead to suboptimal performance. Below, we show how AGT, through its gating mechanism and asynchronous cold-hot gradient updates, severed as an effective strategy for mitigating domain gaps while retaining the benefits of large-scale pre-trained data.

\subsubsection{Revisiting the Domain Adaptation Bound}

Let $\mathcal{D}_S$ denote the source distribution (from which large-scale pretraining data are sampled) and $\mathcal{D}_T$ denote the target distribution (the new hyperspectral dataset). In classical domain adaptation theory~\cite{ben2006analysis}, for any hypothesis $h$ in a hypothesis space $\mathcal{H}$, the error on the target distribution, $\epsilon_T(h)$, can be upper-bounded as:
\begin{equation}
    \label{eq:da_bound}
    \epsilon_T(h) 
    \;\le\; 
    \epsilon_S(h)
    \;+\; 
    \frac{1}{2} \, d_{\mathcal{H}\Delta\mathcal{H}}(\mathcal{D}_S, \mathcal{D}_T)
    \;+\; 
    \lambda,
\end{equation}
where~$\epsilon_S(h)$ is the error on the source distribution,~$d_{\mathcal{H}\Delta\mathcal{H}}(\mathcal{D}_S, \mathcal{D}_T)$ measures the divergence between source and target distributions under $\mathcal{H}$, and~$\lambda$ reflects the best possible joint error across both domains.

In cross-sensor or cross-modality scenarios, the discrepancy $d_{\mathcal{H}\Delta\mathcal{H}}(\mathcal{D}_S, \mathcal{D}_T)$ can be large. Standard fine-tuning faces two major risks:
\begin{itemize}
    \item \textbf{Over-updating:} the pre-trained model with large learning rates, thereby forgetting valuable source-domain knowledge and increasing $\epsilon_S(h)$,
    \item \textbf{Under-updating:} the model with too small learning rate or frozen, failing to adapt to the target domain and keeping $\epsilon_T(h)$ suboptimal.
\end{itemize}

Under this basis, we explain how the proposed AGT design alleviates these issues.

\subsubsection{AGT as an Information Selection Mechanism}

\textit{AGT} introduces a tiny \emph{auxiliary branch} in parallel to the \emph{base model}. We denote: $\theta_{\text{base}}$ as \text{parameters of the pre-trained model}, $\theta_{\text{aux}}$ as \text{parameters of the lightweight branch},
and let $G(\cdot)$ be the cross-attention-based gating module. 
Thus, the final hypothesis $h$ can be expressed as:
\begin{equation}
    \label{eq:hypothesis}
    h(\mathbf{x}; \theta_{\text{base}}, \theta_{\text{aux}})
    \;=\;
    G\bigl(
        f_{\text{base}}(\mathbf{x}; \theta_{\text{base}}),
        \;
        f_{\text{aux}}(\mathbf{x}; \theta_{\text{aux}})
    \bigr),
\end{equation}
where $f_{\text{base}}(\mathbf{x}; \theta_{\text{base}})$ is the feature map from the base model, and 
$f_{\text{aux}}(\mathbf{x}; \theta_{\text{aux}})$ is the feature map from the auxiliary branch.
\begin{itemize}
    \item \textbf{Selective Retention of Pre-trained Knowledge.} Within the gating module, both shallow and deep features from the base model serve as \emph{key-value} pairs, while the auxiliary branch feature forms the \emph{query}. Through cross-attention, the auxiliary branch ``selects'' only the most relevant information from the base model. Irrelevant or conflicting channels are less likely to be activated. Hence, the pre-trained representations learned from large-scale data remain available and are not overwritten, aligning with the core goal of domain adaptation: preserving source knowledge to avoid an excessive increase in $\epsilon_S(h)$.
    \item \textbf{Auxiliary Branch for Target-specific Adaptation.} For target data $\mathcal{D}_T$ that differ significantly from $\mathcal{D}_S$, the auxiliary branch $\theta_{\text{aux}}$ can \emph{freely learn} new spectral-spatial patterns, not fully captured by the base model. By integrating these newly learned features through the gating mechanism in~Eq.\eqref{eq:hypothesis}, the overall hypothesis space effectively \emph{expands} from $\mathcal{H}$ (original base model) to $\widetilde{\mathcal{H}}=\{G(f_{\text{base}},f_{\text{aux}})\}$. This extra flexibility can bridge large domain gaps $d_{\mathcal{H}\Delta\mathcal{H}}(\mathcal{D}_S, \mathcal{D}_T)$. 
From the view of~Eq.\eqref{eq:da_bound}, AGT thus aims to 
(i) keep $\epsilon_S(h)$ low by preserving core parameters and 
(ii) reduce $d_{\mathcal{H}\Delta\mathcal{H}}(\mathcal{D}_S, \mathcal{D}_T)$ with new trainable components $\theta_{\text{aux}}$.
\end{itemize}

\subsubsection{Asynchronous Cold-Hot Updates for Balanced Optimization}

While the above perspective highlights the importance of preserving base-model knowledge, it is equally crucial to adapt efficiently to $\mathcal{D}_T$. AGT uses an \emph{asynchronous cold-hot gradient update} to address this:
\begin{equation}
\begin{aligned}
    \theta_{\text{aux}}^{(t+1)} 
    &= \theta_{\text{aux}}^{(t)} - \alpha \,\nabla_{\theta_{\text{aux}}}\mathcal{L}_{T}\bigl(\theta_{\text{aux}}^{(t)}, \theta_{\text{base}}^{(t)}\bigr),
    \\
    \theta_{\text{base}}^{(t+1)} 
    &= \theta_{\text{base}}^{(t)} - \beta \,\nabla_{\theta_{\text{base}}}\mathcal{L}_{T}\bigl(\theta_{\text{aux}}^{(t)}, \theta_{\text{base}}^{(t)}\bigr),
\end{aligned}
\end{equation}
where $\alpha \gg \beta$. Consequently,
\begin{itemize}
    \item $\theta_{\text{aux}}$ (\textit{hot}) rapidly adapts to domain-specific cues in $\mathcal{D}_T$,
    \item $\theta_{\text{base}}$ (\textit{cold}) is updated more conservatively, preventing catastrophic forgetting of the source knowledge.
\end{itemize}
Hence, this setup finds a middle ground between the extremes of 
(i)~freezing $\theta_{\text{base}}$ (potentially insufficient for large domain shifts), and 
(ii)~fully fine-tuning with large learning rates (risking an overcorrection that erases pre-trained representations).


\subsubsection{Summary of Domain Adaptation Insights}
From a domain adaptation standpoint, the proposed AGT design can be summarized as follows:
\begin{itemize}
    \item \textbf{Gate-based feature fusion} (Eq.~\eqref{eq:hypothesis}) selectively extracts relevant knowledge from the pre-trained base model without inheriting spurious or conflicting information.
    \item \textbf{Auxiliary branch} $\theta_{\text{aux}}$ captures new spectral-spatial patterns crucial for the target hyperspectral domain, effectively reducing the distribution gap $d_{\mathcal{H}\Delta\mathcal{H}}(\mathcal{D}_S,\mathcal{D}_T)$.
    \item \textbf{Asynchronous cold-hot updates} maintain a stable optimization path that balances preserving source knowledge and adapting to the new domain.
\end{itemize}
As validated by our experiments in Section~IV, these ideas collectively enable robust and efficient adaptation, even under challenging cross-sensor or cross-modality conditions where domain discrepancies are pronounced.

\section{Experiments}

\subsection{Data Description}
In order to comprehensively evaluate the performance of our proposed method under diverse conditions, we selected six representative datasets. These datasets vary in sensor type, spectral band count, spatial resolution, and scene characteristics. 
This selection addresses three key challenges: (1) scene specificity (differing land-cover types and background complexities), (2) spatial resolution differences (mixed pixels in low resolution versus fine spatial details in high resolution), and (3) the challenge of effectively fusing spectral and spatial information under varying conditions. Overall, the diversity of the chosen datasets ensures a fair and rigorous evaluation of the proposed method across different imaging scenarios.  The sample distribution information is listed in Table ~\ref{tab:dataset-1} and Table~\ref{tab:dataset-2}, and the detail introduction of these datasets are as follows: 


\textbf{Indian Pines} dataset was acquired by the Airborne Visible/Infrared Imaging Spectrometer (AVIRIS) sensor during a flight campaign over the Indian Pines agricultural site in Indiana, USA. The Indian Pines dataset consists of a HSI with a high spectral resolution, capturing information in 224 contiguous spectral bands covering the wavelength range from 0.4 µm to 2.5 µm. The spatial resolution of the dataset is 145×145 pixels, making it relatively small compared to some other hyperspectral datasets. The primary objective of the Indian Pines dataset is to classify the various land-cover objects present in the agricultural area, which includes crops, bare soil, and different types of vegetation. The dataset contains 16 different land-cover classes in total.

\textbf{Pavia University} dataset was collected by the Reflective Optics System Imaging Spectrometer (ROSIS) sensor during a flight campaign over Pavia, a city in northern Italy. The dataset consists of a HSI with a high spectral resolution and geometric resolution of 1.3 meters. It captures information in 103 contiguous spectral bands covering the wavelength range from 0.43 µm to 0.86 µm. The spatial resolution of the Pavia University dataset is 610×340 pixels, making it of moderate size. The dataset contains 9 land-cover classes, representing various ground objects and materials present in the urban and agricultural regions of Pavia. The main objective of the Pavia University dataset is to classify the different land-cover classes present in the scene, which includes urban areas, vegetation, soil, and other materials.

\begin{table*}[!t]
\setlength{\abovecaptionskip}{0.cm}
\setlength{\belowcaptionskip}{-0.cm}
\renewcommand{\arraystretch}{1.1}
\caption{Sample distribution information of Indian Pines \& Pavia University Houston \& University Datasets}
\label{tab:dataset-1}
\setlength{\tabcolsep}{10pt}
\centering
\begin{tabular}{c|cc|cc|cc}
\hline
\multicolumn{3}{c}{Indian Pines} & \multicolumn{2}{|c}{Pavia University} & \multicolumn{2}{|c}{Houston University} \\
\hline
Class & Land Cover Type & No.of Samples & Land Cover Type & No.of Samples & Land Cover Type & No.of Samples \\
\hline
1	& Alfalfa       & 46  & Asphalt      & 6631    & Healthy Grass   &  1251 \\  
2	& Corn-notill   & 1428  & Meadows    & 18649  & Stressed Grass   &  1254 \\
3	& Corn-mintill  & 830 & Gravel & 2099    & Synthetic Grass &  697  \\
4	& Corn          & 237 & Trees &  3064    & Trees           &  1244 \\
5	& Grass-pasture &  483  &Painted Metal Sheets            &1345   & Soil            &  1242 \\
6	& Grass-trees          & 730  &  Bare Soil               &5029   & Water           &  325  \\
7	& Grass-pasture-mowed	           & 28  &Bitumen              &  1330  & Residential     &  1268 \\
8   & Hay-windrowed &  478 & Self-Blocking Bricks              &3682   & Commercial      &  1244 \\
9	& Oats              & 20   &Shadows            & 947    & Road            &  1252 \\
10  & Soybean-notill                  &   972    &  -                   &   -   & Highway         &  1227 \\
11  & Soybean-mintill                   &   2455    &  -                   &   -   & Railway         &  1235 \\
12  & Soybean-clean                   &   593   &  -                   &   -   & Parking Lot 1   &  1233 \\
13  & Wheat                   &   205   &  -                   &   -   & Parking Lot 2   &  469  \\
14  & Woods                  &   1265   &  -                   &   -   & Tennis Court    &  428  \\
15  & Buildings-Grass-Trees-Drives   &   386    &  -                   &   -   & Running Track   &  660  \\
16  &  Stone-Steel-Towers    &   93    &  -                   &   -   & -   &  - \\
\hline
    & Total                & 10249  & Total                & 42776   & Total           & 15029 \\
\hline
\end{tabular}
\end{table*}

\begin{table*}[!t]
\setlength{\abovecaptionskip}{0.cm}
\setlength{\belowcaptionskip}{-0.cm}
\renewcommand{\arraystretch}{1.1}
\caption{Sample distribution information of WHU-Hi Datasets}

\label{tab:dataset-2}
\setlength{\tabcolsep}{10pt}
\centering
\begin{tabular}{c|cc|cc|cc}
\hline
\multicolumn{3}{c}{WHU-Hi-LongKou} & \multicolumn{2}{|c}{WHU-Hi-HanChuan} & \multicolumn{2}{|c}{WHU-Hi-HongHu} \\
\hline
Class & Land Cover Type & No.of Samples & Land Cover Type & No.of Samples & Land Cover Type & No.of Samples \\
\hline
1	& Corn       & 34511  & Strawberry      & 44735    & Red roof   &  14041 \\  
2	& Cotton   & 8374  & Cowpea    & 22753 & Road   &  3512 \\
3	& Sesame  & 3031 & Soybean & 10287    & Bare soil & 21821  \\
4	& Broad-leaf soybean          & 5353 & Sorghum	 &  3064    & Cotton          &  163285 \\
5	& Narrow-leaf soybean &  4151  &Water spinach            &1200   &Cotton firewood            & 6218 \\
6	& Rice          & 11854  &  Watermelon	               &4533   & Rape           &  44557  \\
7	& Water	           & 67056  &Greens              &  5903  & Chinese cabbage     & 24103\\
8   & Roads and houses &  7124 & Trees              &17978   & Pakchoi      &  4054 \\
9	& Mixed weed              & 5229   &Grass            & 9469    & Cabbage           &  10819 \\
10  &   -                &  -     &  Red roof                   &   10516   & Tuber mustard         &  12394\\
11  &   -                 &  -    &  Gray roof                  &   16911  &Brassica parachinensis         &  11015 \\
12  &   -                 &  -   &  Plastic	                  &   3679   & Brassica chinensis   & 8954 \\
13  &  -                  &  -   &  Bare soil                   &   9116	 & Small Brassica chinensis   &  22507	  \\
14  &  -               & -     &  Road                 &  18560  &Lactuca sativa    &  7356  \\
15  &  -  &  -    &  Bright object                   &   1136   &Celtuce   &  1002  \\
16  & -     &  -     & Water                   &   75401   & Film covered lettuce &  7262 \\
17  & -     &  -     & -                   &  -   & Romaine lettuce &  3010 \\
18  & -     &  -     & -                   &  -   & Carrot & 3217  \\
19 & -     & -      & -                   &  -   & White radish &  8712 \\
20  & -     &  -     & -                   &  -   & Garlic sprout	 & 3486 \\
21  & -     &  -     & -                   &  -   & Broad bean	 & 1328  \\
22  & -     &  -     & -                   &  -   & Tree	 &  4040 \\
\hline
    & Total                & 146683  & Total                & 255241   & Total           & 381693 \\
\hline
\end{tabular}
\end{table*}

\textbf{Houston University} dataset was captured by the ITRES-CASI 1500 hyperspectral imager over the University of Houston campus and the neighboring urban area in June 2012. The dataset comprises a HSI with a high spectral resolution, covering 144 contiguous spectral bands that span the wavelength range from 0.36 µm to 1.05 µm. With a spatial resolution of 349×1905 pixels, the Houston University dataset is relatively large, providing a detailed representation of the urban and suburban regions in the vicinity of the university. Houston University dataset consists of 349 × 1905 pixels. The dataset encompasses 15 distinct land-cover classes, including urban structures, roads, vegetation, water bodies, and other land-use categories typically encountered in urban environments.


\textbf{WHU-Hi-LongKou} dataset~\cite{zhong2020whu} was captured on July 17, 2018, in Longkou Town, Hubei province, China. The imaging was conducted using a DJI Matrice 600 Pro UAV, equipped with an 8-mm focal length Headwall Nano-Hyperspec sensor, flying at an altitude of 500 meters. The resulting dataset consists of images with dimensions of 550 × 400 pixels and a spatial resolution of about 0.463 meters. It spans 270 spectral bands, covering wavelengths from 400 to 1000 nm. The scene mainly includes agricultural fields with six types of crops: corn, cotton, sesame, broad-leaf soybean, narrow-leaf soybean, and rice.

\textbf{WHU-Hi-HanChuan} dataset~\cite{zhong2020whu} was gathered on June 17, 2016 in Hanchuan, Hubei province, China. The UAV platform used was a Leica Aibot X6, featuring a 17-mm focal length Headwall Nano-Hyperspec sensor, operating at an altitude of 250 meters. The dataset comprises images with a size of 1217 × 303 pixels and a spatial resolution of approximately 0.109 meters, spanning 274 spectral bands from 400 to 1000 nm. The study area, a transitional zone between rural and urban land, includes buildings, water bodies, and fields with seven different crops: strawberry, cowpea, soybean, sorghum, water spinach, watermelon, and greens.

\textbf{WHU-Hi-HongHu} dataset~\cite{zhong2020whu} was collected on November 20, 2017 in Honghu City, Hubei province, China. The data acquisition was carried out using a DJI Matrice 600 Pro UAV equipped with a 17-mm focal length Headwall Nano-Hyperspec sensor. The study area is agriculturally diverse, featuring multiple crop types and various cultivars of the same crop, such as Chinese cabbage and cabbage, as well as different varieties of Brassica chinensis. The UAV operated at a flight altitude of 100 meters, capturing imagery with dimensions of 940 × 475 pixels, across 270 spectral bands ranging from 400 to 1000 nm. The spatial resolution of the images is approximately 0.043 meters.

\subsection{Experiment Design}

The experimental setup encompasses two comparative segments for classification evaluation. The first part involves basic model comparison experiments, while the second part focuses on comparing various tuning strategies.

In the basic model comparison part, we partition each target HSI dataset into two subsets: the training set and test set. We randomly extract 150 samples from each category as training samples and take the rest as the test samples in the Pavia University, Houston University and WHU-Hi datasets. For the Indian Pines dataset, we sample a small number of pixels from classes with scarce samples to ensure a more balanced representation for the experiments. Concretely, we randomly select 10 samples from each of the following six classes: Alfalfa (class 1), Grass/trees (class 5), Grass/pasture-mowed (class 7), Oats (class 9), Buildings-grass-trees (class 15), and Stone-steel towers (class 16). For the remaining classes, we sample 150 pixels from each class.  In this Section IV.D, we compare our proposed Tri-Former model with other state-of-the-art algorithms. 

In addition, we compare the proposed AGT approach with other popular tuning methods on HSIs collected from different sensors. The tuning strategies comparison results are presented in Section IV.G and Section IV.H. These experiments involve data from two source sensors: Salinas (captured by AVIRIS) and Pavia Center (captured by ROSIS). For the model trained on the data collected by one sensor, we fine-tune it on the target data collected by another sensor with a limited number of samples. We design different experimental groups, each containing 25, 50, and 75 labeled pixels per class, respectively. Furthermore, in Section IV.I, we evaluate the proposed method by cross-modality fine tuning experiments. Specifically, the basic Tri-Former model pretrained on RGB images is fintuned on different HSI datasets with different tuning method.

Through comprehensive and systematic experiments, we demonstrate the effectiveness of the proposed method in addressing HSI classification. To ensure the fairness and stability of the comparison, we repeat each experiment five times and take the average values as the final results.

\begin{figure}
	\begin{center}
		\includegraphics[width=1.0 \linewidth]{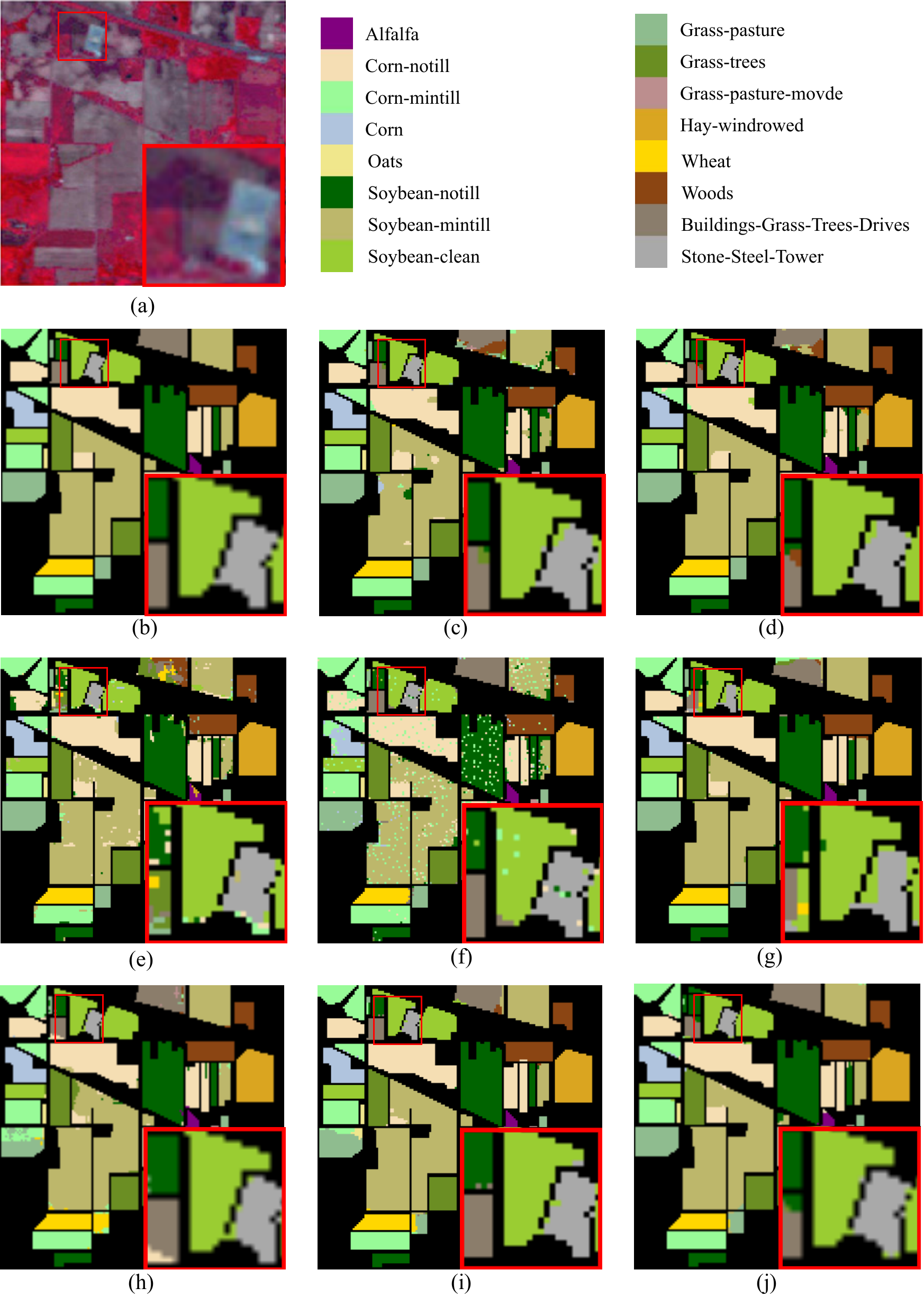}
	\end{center}
	\vspace{-0.3cm}
	\caption{Results on Indian Pines. (a) Composite color map; (b) Ground truth; (c) 3D-CNN; (d) LWNet; (e) SpectralFormer; (f) GraphGST; (g) SSFTT; (h) MHCFormer; (i) MASSFormer; (j) Tri-Former.}
	\label{fig:indian}
\end{figure}

\subsection{Implementation Details}

The optimization of the proposed method consists of two stages. The entire learning process is conducted on the server with 4 NVIDIA 3090 GPUs. During training, following the experience of previous works such as 3D-CNN, LWNet, etc. the training samples are cropped as $27 \times 27$ spatial resolution patches, with a batch size of 96 on all three training dataset. We utilize the AdamW optimizer with a learning rate and weight decay value of $1 \times 10^{-3}$ and $1 \times 10^{-5}$, respectively. Following the training strategy proposed in ConvNext, the learning rate decays from $1 \times 10^{-3}$ to $1 \times 10^{-6}$ following the cosine scheduler. The warming-up stage comprises 5 epochs and all the models are trained for 300 epochs. To further tune the final network, we crop patches with spatial resolutions of $13 \times 13$ and $27 \times 27$, which are fed into the auxiliary branch and basic model, respectively. Batch sizes are set to 12 for all datasets during tuning. 

\begin{table}[!t]
\setlength{\abovecaptionskip}{0.cm}
\setlength{\belowcaptionskip}{-0.cm}
\renewcommand{\arraystretch}{1.2}
\caption{Compared Methods Overview}
\label{tab:method_overview}
\centering
\begin{threeparttable}
\resizebox{\linewidth}{!}{
\begin{tabular}{c|ccccccc}
\hline
 Method                   & 3D-CNN\footnote{\url{https://github.com/eecn/Hyperspectral-Classification}}~\cite{chen2016deep}  & LWNet\footnote{\url{https://github.com/hkzhang91/LWNet}}~\cite{2019Hyperspectral}   & SpectralFormer\footnote{\url{https://github.com/danfenghong/IEEE_TGRS_SpectralFormer}}~\cite{hong2021spectralformer} & SSFTT\footnote{\url{https://github.com/zgr6010/HSI_SSFTT}}~\cite{sun2022spectral}   & GraphGST\footnote{\url{https://github.com/yuanchaosu/TGRS-graphGST}}~\cite{gst} & MHCFormer\footnote{\url{https://github.com/Tikiten/MHCFormer}}~\cite{shi2023mhcformer}     & MassFormer\footnote{\url{https://github.com/hz63/ MASSFormer}}~\cite{10506482} \\
\hline
Year                & TGRS16  & TGRS19  & TGRS21         & TGRS22  & TGRS24   & TIE24  & TGRS24     \\
Architecture        & ConvNet & ConvNet & Transformer    & Transformer & Transformer & Transformer & Transformer \\
\hline
\end{tabular}
}
\end{threeparttable}
\end{table}

\begin{table}[!t]
\setlength{\abovecaptionskip}{0.cm}
\setlength{\belowcaptionskip}{-0.cm}
\renewcommand{\arraystretch}{1.1}
\caption{Comparison Experimental Results on Indian Pines Using 150 Training Samples Each Class}
\label{tab:Indian_150}
\centering
\begin{threeparttable}
\resizebox{\linewidth}{!}{
\begin{tabular}{c|ccccccc|c}
\hline
Models & 3D-CNN & LWNet & SpectralFormer & GraphGST & SSFTT  & MHCFormer & MASSFormer & Tri-Former \\
\hline
    1  & 91.67   & 97.22 & 77.78  & \textbf{100.00} & \textbf{100.00} & \textbf{100.00}   & \textbf{100.00}   & 97.14 \\
    2  & 94.84   & 96.01 & 90.06  & 94.91   & 94.99   & 94.60  & 96.09 & \textbf{99.45} \\
    3  & 99.12   & 98.82 & 94.26  & 99.26   & 99.12   & 98.97 & 99.56 & \textbf{99.71} \\
    4  & \textbf{100.00} & \textbf{100.00} & 98.85  & 74.89   & \textbf{100.00} & \textbf{100.00}   & \textbf{100.00}   & \textbf{100.00} \\
    5  & \textbf{100.00} & \textbf{100.00} & 99.10  & 96.99   & \textbf{100.00} & 57.08 & 78.86 & 91.45 \\
    6  & \textbf{100.00} & 99.83  & 98.97  & 97.76   & 99.83   & 100   & 99.83 & \textbf{100.00} \\
    7  & \textbf{100.00} & \textbf{100.00} & 94.44  & 94.44   & \textbf{100.00} & \textbf{100.00}   & \textbf{100.00}   & \textbf{100.00} \\
    8  & \textbf{100.00} & \textbf{100.00} & \textbf{100.00}  & \textbf{100.00} & \textbf{100.00} & \textbf{100.00}   & 99.70  & \textbf{100.00} \\
    9  & \textbf{100.00} & \textbf{100.00} & \textbf{100.00}  & \textbf{100.00} & \textbf{100.00} & 99.15 & \textbf{100.00}   & \textbf{100.00} \\
    10 & 99.39   & \textbf{100.00} & 96.35  & 99.15   & 99.51   & 96.79 & 99.76 & 99.51 \\
    11 & 92.41   & \textbf{98.39} & 90.54  & 93.02   & 95.05   & 98.87 & 99.74 & 98.13 \\
    12 & 98.42   & \textbf{99.77} & 96.61  & 96.61   & 99.55   & 100   & 99.32 & 99.56 \\
    13 & \textbf{100.00} & \textbf{100.00} & \textbf{100.00}  & 97.95   & \textbf{100.00} & \textbf{100.00}   & \textbf{100.00}   & \textbf{100.00} \\
    14 & 98.92   & 99.91  & 98.92  & 99.46   & 98.82   & \textbf{100}   & 99.64 & \textbf{100.00} \\
    15 & 66.49   & 70.74  & 68.35  & \textbf{100.00} & 89.63   & 90.69 & 88.56 & 91.76 \\
    16 & 90.36   & 91.57  & 84.33  & \textbf{97.59} & 83.13   & 96.39 & 92.77 & 89.16 \\
\hline
    OA & 95.23   & 97.46  & 93.08  & 96.02   & 97.12   & 95.36 & 97.48 & \textbf{98.41} \\
    AA & 95.73   & 97.02  & 93.04  & 96.38   & 97.54   & 95.78 & 97.11 & \textbf{97.87} \\
    K  & 94.47   & 97.05  & 91.97  & 95.41   & 96.66   & 94.63 & 97.07 & \textbf{98.17} \\
\hline
\end{tabular}
}
\end{threeparttable}
\end{table}

\begin{table}[!t]
\setlength{\abovecaptionskip}{0.cm}
\setlength{\belowcaptionskip}{-0.cm}
\renewcommand{\arraystretch}{1.05}
\caption{Comparison Experimental Results on Pavia University Using 150 Training Samples Each Class}
\label{tab: PaviaU_150}
\centering
\begin{threeparttable}
\resizebox{\linewidth}{!}{
\begin{tabular}{c|ccccccc|c}
\hline
Models & 3D-CNN & LWNet & SpectralFormer & GraphGST   & SSFTT  &MHCFormer	&MASSFormer & Tri-Former \\
\hline
    1 & 91.09 & 96.06 & 83.13            & 98.06 &94.69  &96.48	&99.24 & \textbf{99.98} \\
    2 & 92.05 & 95.69 & 92.83            & 98.03 &98.74 &98.90	&99.76 & \textbf{99.91} \\ 
    3 & 92.45 & 95.95 & 89.34            & 92.82 &98.50  &98.97	&\textbf{100.00}  & \textbf{100.00} \\ 
    4 & 98.55 & 98.68 & 97.60            & 99.11 &95.05 &92.28 &99.07 & \textbf{99.18}\\ 
    5 & 99.52 & 99.92 & \textbf{100.00}           & \textbf{100.00} & 97.43  &99.58	&\textbf{100.00} & \textbf{100.00} \\ 
    6 & 93.14 & 98.09 & 90.32            & 90.92 &\textbf{99.96} &99.51	&\textbf{100.00} & \textbf{100.00} \\ 
    7 & 97.48 & 99.76 & 96.99            & \textbf{100.00} & \textbf{100.00}  &\textbf{100.00}	&\textbf{100.00} & 99.92 \\ 
    8 & 95.20 & 99.11 & 92.63            & 95.95 &99.08  &97.11	&\textbf{99.97} & 99.77 \\ 
    9 & 97.28 & 98.82 & 99.65            & 99.74 &99.88   &98.49	&\textbf{100.00} & \textbf{100.00}\\ 
    \hline
    OA & 93.26 & 96.86 & 91.65           & 98.06 &98.03  &98.02 &99.71 & \textbf{99.88}\\
    AA & 95.20 & 98.01 & 93.61           & 98.18 &98.15  &97.93	&99.78 & \textbf{99.86}\\ 
    K & 91.13 & 95.84 & 89.00            & 97.41 &97.38   &97.36	&99.61 & \textbf{99.83}\\ 
\hline
\end{tabular}
}
\end{threeparttable}
\end{table}

\subsection{Comparison with state-of-the-art methods}

In this section, we conduct a comparative analysis of our proposed Tri-Former against two ConvNet-based and five transformer-based HSI classification methods. The detail information about comparison methods are lists in Table~\ref{tab:method_overview}. Among them, ConvNet-based methods (e.g., 3D-CNN and LWNet) use fixed convolutional filters to extract spectral–spatial features but lack the adaptive long-range dependency modeling provided by transformers. The remaining methods are Transformer-based methods. Although these methods incorporate self-attention, they do not fully leverage the unique characteristics of HSIs. However, Our Tri-Former adopts a triplet structure (spectral, spatial, and 3D convolution branches) and uses an attention-gated tuning strategy (AGT) to bridge the domain gap, allowing cross-modal (RGB) pretraining.

\begin{figure}
	\begin{center}
		\includegraphics[width=0.95 \linewidth]{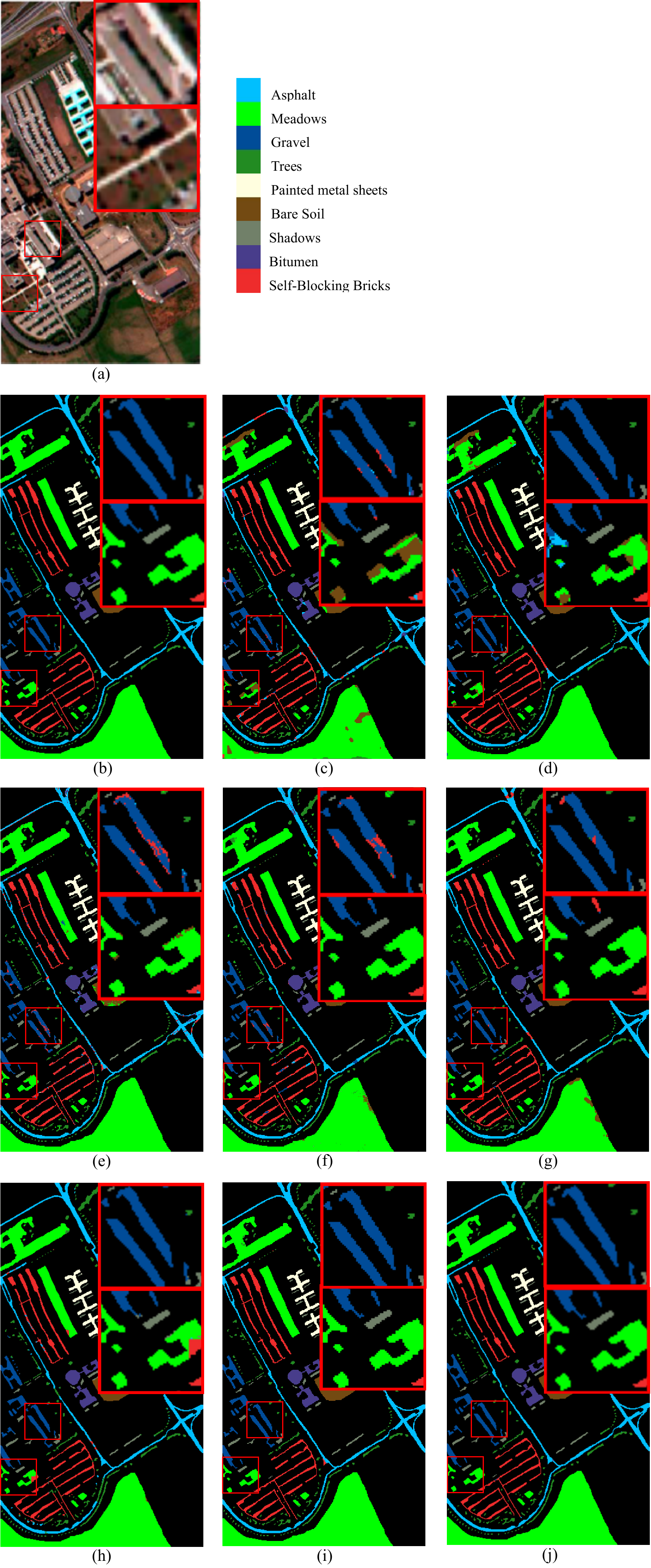}
	\end{center}
	\vspace{-0.5cm}
	\caption{Results on Paiva University. (a) Composite color map; (b) Ground truth; (c) 3D-CNN; (d) LWNet; (e) SpectralFormer; (f) GraphGST; (g) SSFTT;  (h) MHCFormer; (i) MASSFormer; (j) Tri-Former.}
	\label{fig:paviau}
\end{figure}

\begin{figure}
	\begin{center}
		\includegraphics[width=1.0 \linewidth]{ 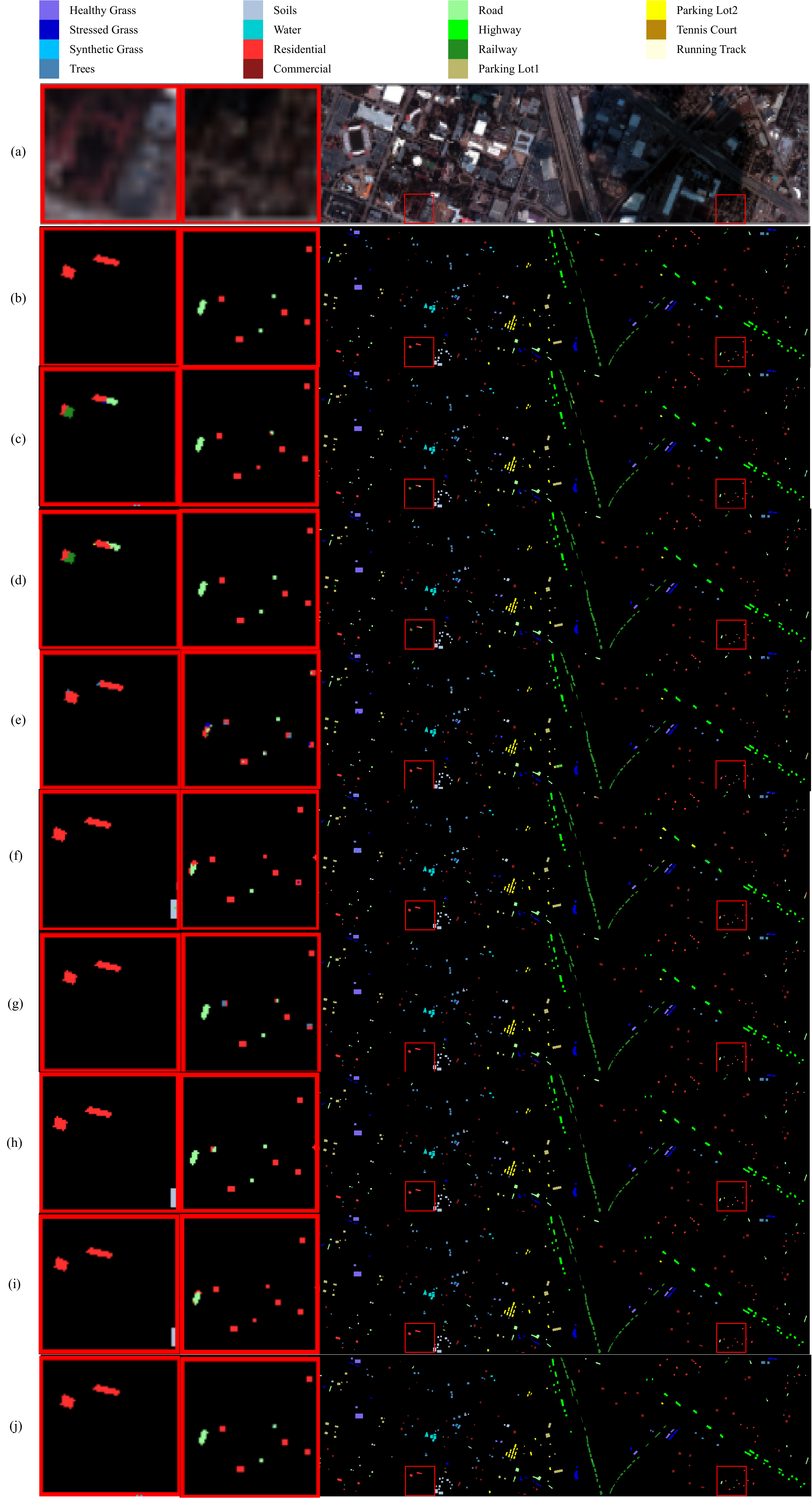}
	\end{center}
	\vspace{-0.3cm}
	\caption{Results on Houston University.  (a) Composite color map; (b) Ground truth; (c) 3D-CNN; (d) LWNet; (e) SpectralFormer; (f) GraphGST; (g) SSFTT; (h) MHCFormer; (i) MASSFormer; (j) Tri-Former.}
	\label{fig:hu}
\end{figure}

\begin{figure}
	\begin{center}
		\includegraphics[width=1.0 \linewidth]{ 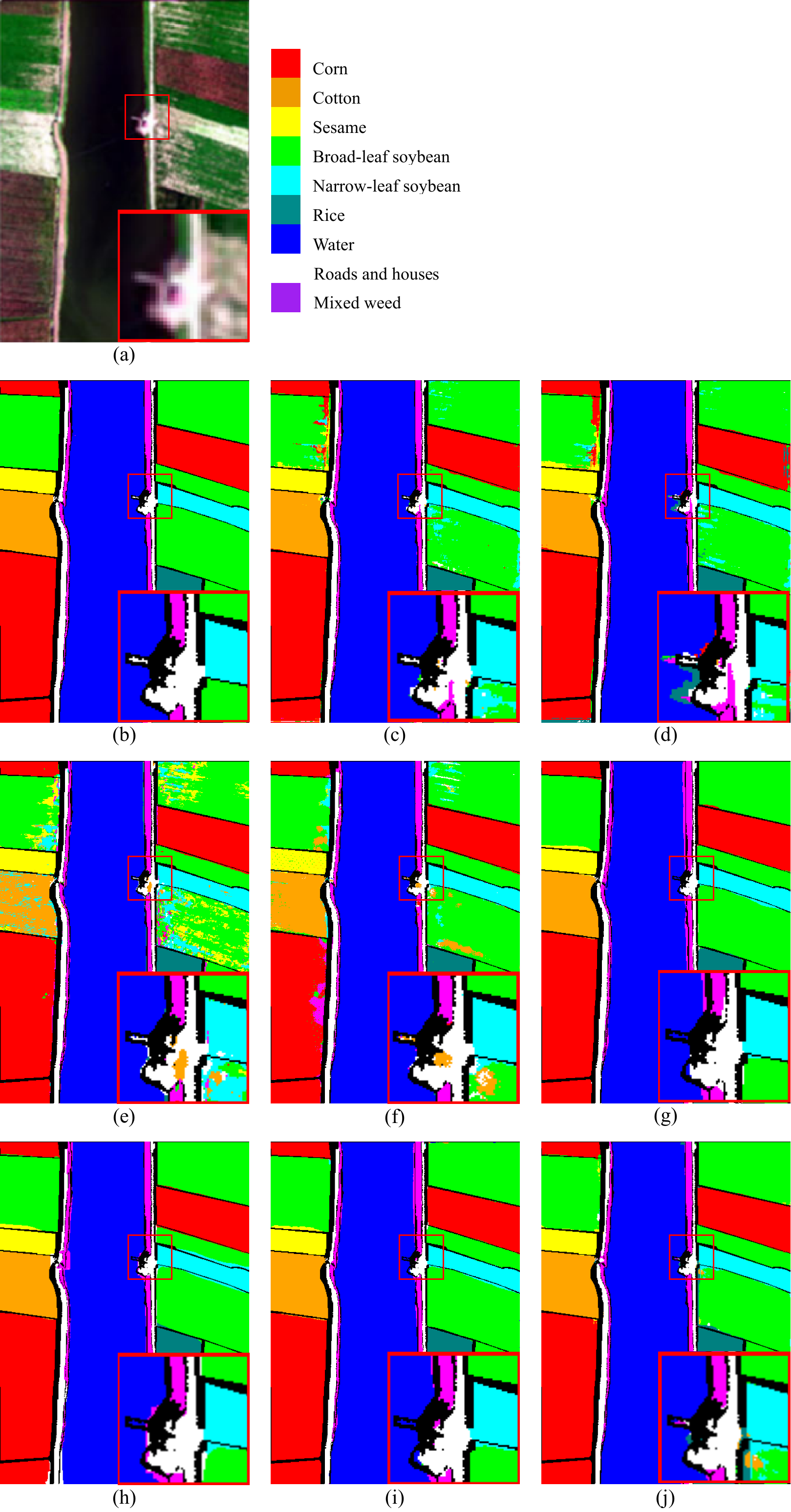}
	\end{center}
	\vspace{-0.3cm}
	\caption{Results on WHU-Hi-LongKou.  (a) Composite color map; (b) Ground truth; (c) 3D-CNN; (d) LWNet; (e) SpectralFormer; (f) GraphGST; (g) SSFTT; (h) MHCFormer; (i) MASSFormer; (j) Tri-Former.}
	\label{fig:WHU-Hi-LongKou}
\end{figure}

\begin{figure}
	\begin{center}
		\includegraphics[width=1.0 \linewidth]{ 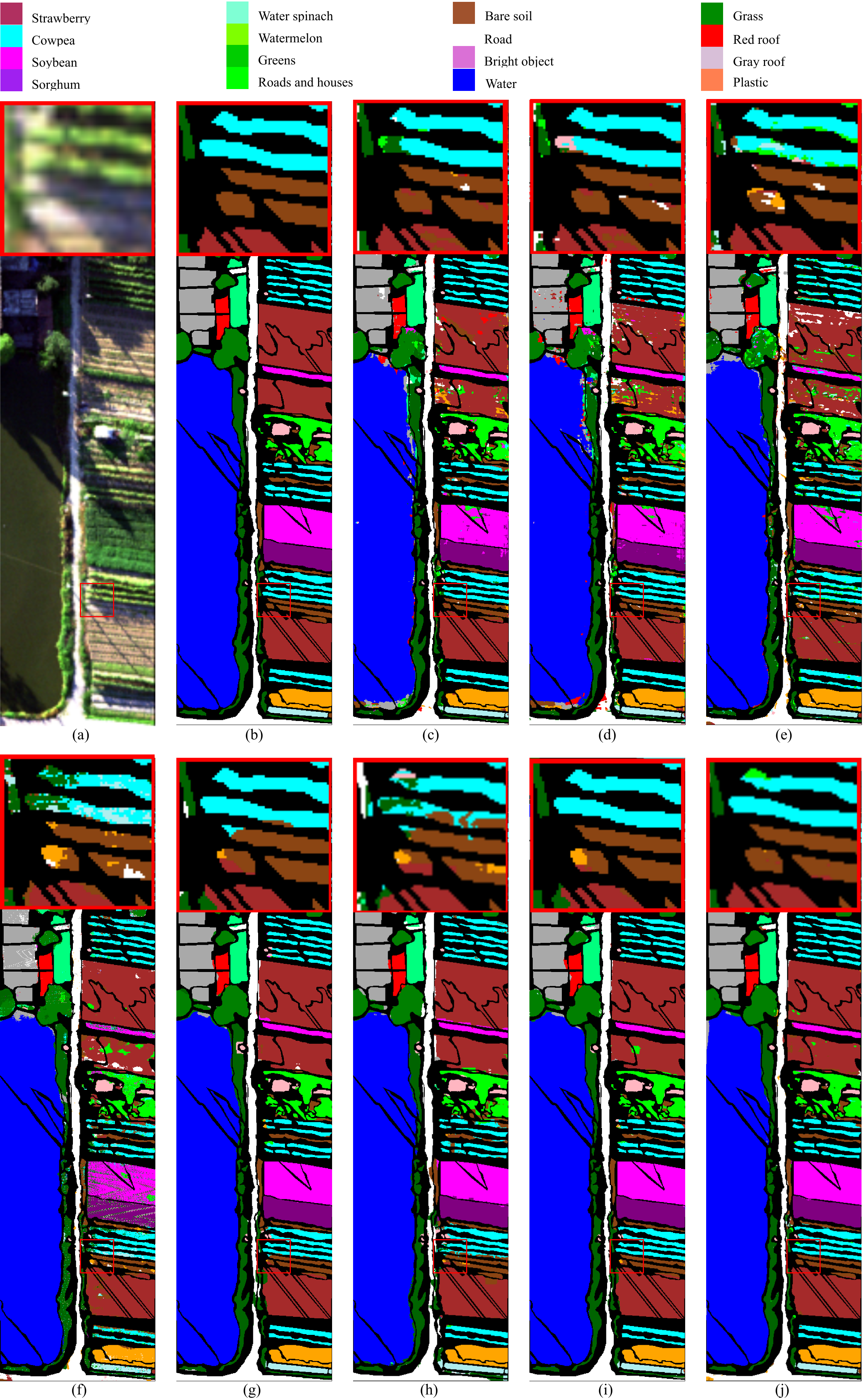}
	\end{center}
	\vspace{-0.3cm}
	\caption{Results on WHU-Hi-HanChuan
    .  (a) Composite color map; (b) Ground truth; (c) 3D-CNN; (d) LWNet; (e) SpectralFormer; (f) GraphGST; (g) SSFTT; (h) MHCFormer; (i) MASSFormer; (j) Tri-Former.}
	\label{fig:WHU-Hi-HanChuan}
\end{figure}

\begin{figure}
	\begin{center}
		\includegraphics[width=1.0 \linewidth]{ 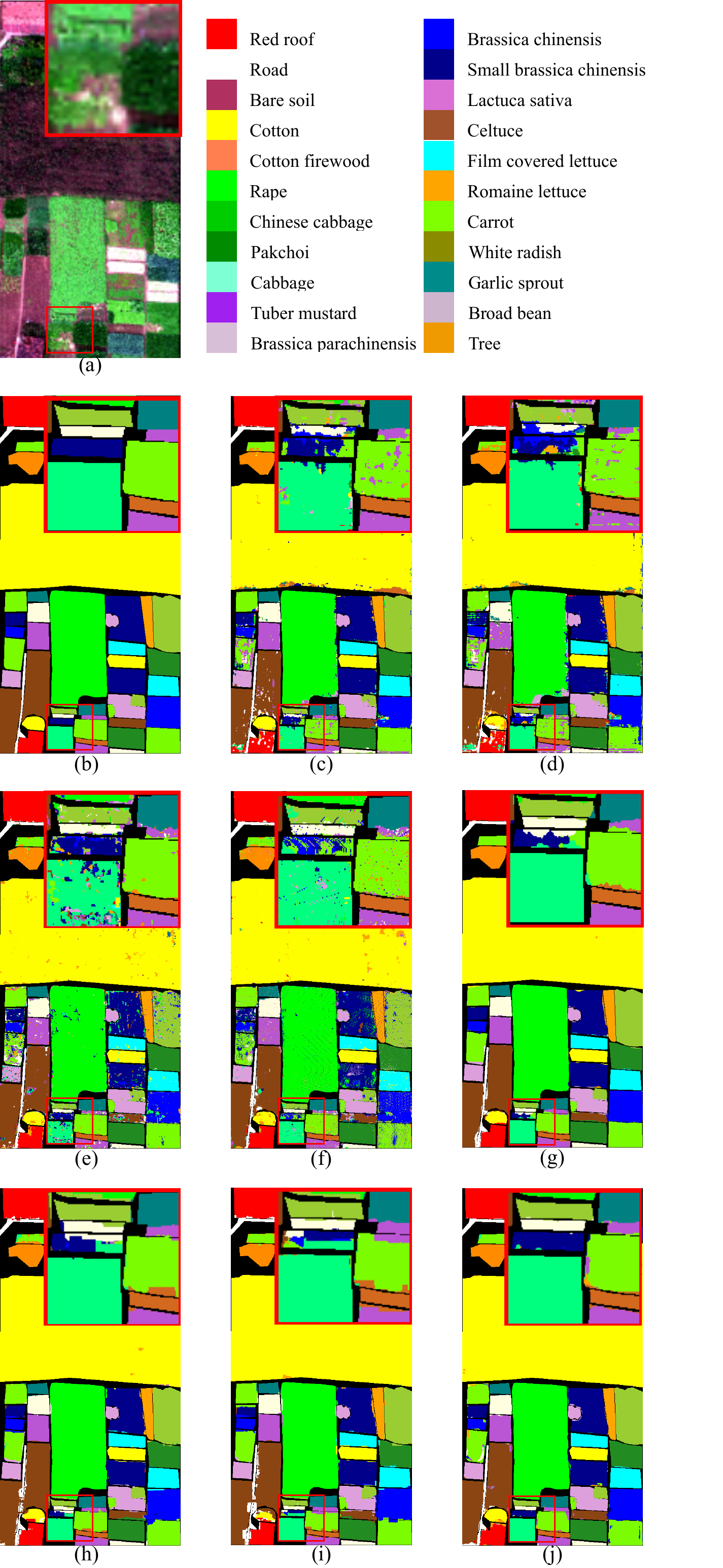}
	\end{center}
	\vspace{-0.3cm}
	\caption{Results on WHU-Hi-HongHu. ( (a) Composite color map; (b) Ground truth; (c) 3D-CNN; (d) LWNet; (e) SpectralFormer; (f) GraphGST; (g) SSFTT; (h) MHCFormer; (i) MASSFormer; (j) Tri-Former.}
	\label{fig:WHU-Hi-HongHu}
\end{figure}

The experimental results are presented in Table~\ref{tab:Indian_150} to Table~\ref{tab:honghu_150}, and the visual comparisons are shown in Fig.~\ref{fig:indian} to Fig.~\ref{fig:WHU-Hi-HongHu}. 

The quantitative analysis results of the aforementioned approaches on the Indian Pines dataset are listed in Table~\ref{tab:Indian_150}. Table~\ref{tab:Indian_150} shows that the proposed Tri-Former achieves the highest OA of 98.41\%, outperforming other ConvNet-based HSI classification methods like 3D-CNN (95.23\%), LWNet (97.46\%),  SpectralFormer (93.08\%), GraphGST (96.02\%) and SSFTT (97.12\%). Among them, SSFTT incorporation of convolutional operations within the transformer structure, which benefits its performance. This improvement further substantiates some of the points we have mentioned in this work. Introducing convolutional operations to construct hybrid transformer models for processing HSI is a more suitable approach. The proposed Tri-Former further improved the OA metric of SSFTT by 1.29\%, demonstrating its remarkable capability to accurately classify HSIs in this challenging dataset.

\begin{table}
\setlength{\abovecaptionskip}{0.cm}
\setlength{\belowcaptionskip}{-0.cm}
\renewcommand{\arraystretch}{1.05}
\caption{Comparison Experimental Results on Houston University Using 150 Training Samples Each Class}
\label{tab: HoustonU_150}
\centering
\begin{threeparttable}
\resizebox{\linewidth}{!}{
\begin{tabular}{c|ccccccc|c}
\hline
Models & 3D-CNN & LWNet & SpectralFormer & GraphGST & SSFTT  & MHCFormer  & MASSFormer & Tri-Former \\
\hline
1      & 96.64  & 98.64  & 96.00          & 78.47   & 98.00  & \textbf{100.00}   & \textbf{100.00}   & 99.64 \\
2      & 95.20  & 99.01  & 89.31          & 83.51   & 97.10  & 95.47 & \textbf{99.73} & 97.92 \\
3      & 99.82  & \textbf{100.00} & 99.09          & 84.82   & 99.09  & \textbf{100.00}   & \textbf{100.00}   & \textbf{100.00} \\
4      & 97.62  & 99.54  & 86.47          & 79.43   & 97.90  & 97.90  & \textbf{99.91} & 99.73 \\
5      & 99.45  & \textbf{100.00} & 95.88          & 97.43   & \textbf{100.00} & \textbf{100.00}   & \textbf{100.00}   & 99.91 \\
6      & \textbf{100.00} & \textbf{100.00} & 86.86          & 88.57   & \textbf{100.00}   & \textbf{100.00}   & \textbf{100.00}   & \textbf{100.00} \\
7      & 91.68  & 96.06  & 73.23          & 62.88   & 98.66  & 95.01 & 98.93 & \textbf{99.11} \\
8      & 94.70  & 96.71  & 85.56          & 54.39   & 97.62  & 96.07 & \textbf{100.00}   & 97.17 \\
9      & 96.82  & 97.19  & 85.57          & 54.38   & \textbf{99.82}  & 96.82 & 99.36 & 98.82 \\
10     & 98.33  & 99.44  & 91.55          & 70.10   & \textbf{100.00}   & 99.81 & \textbf{99.91} & 99.63 \\
11     & 97.05  & \textbf{100.00} & 92.35          & 51.34   & \textbf{100.00}   & \textbf{100.00}   & 99.72 & 97.05 \\
12     & 96.12  & 98.25  & 86.98          & 60.85   & 96.21  & 98.43 & 99.63 & \textbf{100.00}   \\
13     & 95.61  & \textbf{100.00} & 82.45          & 86.20   & 99.37  & \textbf{100.00}   & \textbf{100.00}   & \textbf{100.00} \\
14     & \textbf{100.00}   & \textbf{100.00}   & \textbf{100.00}   & 94.96   & \textbf{100.00}   & \textbf{100.00}   & \textbf{100.00}   & \textbf{100.00} \\
15     & \textbf{100.00}   & \textbf{100.00}   & 97.65          & 86.67   & \textbf{100.00}   & 99.8  & \textbf{100.00}   & \textbf{100.00} \\
\hline
OA     & 96.58  & 98.70  & 89.19          & 94.42   & 98.69  & 98.23 & \textbf{99.76} & 99.05 \\
AA     & 97.14  & 98.99  & 89.93          & 95.13   & 98.92  & 98.63 & \textbf{99.81} & 99.26 \\
K      & 96.29  & 98.59  & 88.28          & 93.94   & 98.57  & 98.08 & \textbf{99.74} & 98.97 \\
\hline
\end{tabular}
}
\end{threeparttable}
\end{table}

\begin{table}[!t]
\setlength{\abovecaptionskip}{0.cm}
\setlength{\belowcaptionskip}{-0.cm}
\renewcommand{\arraystretch}{1.1}
\caption{Comparison Experimental Results on WHU-Hi-Longkou Using 150 Training Samples Each Class}
\label{tab:longkou_150}
\centering
\begin{threeparttable}
\resizebox{\linewidth}{!}{
\begin{tabular}{c|ccccccc|c}
\hline
Models & 3D-CNN & LWNet & SpectralFormer & GraphGST & SSFTT  & MHCFormer  & MASSFormer & Tri-Former \\
\hline
1  & 99.39  & 97.28  & 98.97  & \textbf{99.87}  & 96.21  & 99.47  & 99.67  & 99.74 \\
2  & 98.39  & 98.12  & 86.02  & 99.67  & 95.39  & 99.27  & \textbf{99.98}  & 99.87  \\
3  & 99.86  & \textbf{100.00} & 99.27  & \textbf{100.00} & 99.44  & 99.60  & \textbf{100.00} & \textbf{100.00} \\
4  & 94.22  & 93.80  & 81.32  & 98.62  & 94.00  & 97.42  & \textbf{98.78}  & 98.26  \\
5  & \textbf{100.00} & \textbf{100.00} & 98.17  & 99.80  & 99.65  & \textbf{100.00} & 99.85  & 99.38  \\
6  & 99.39  & 97.56  & 98.73  & 98.09  & 99.03  & 99.22  & 99.69  & \textbf{100.00} \\
7  & 99.82  & 99.20  & 99.05  & 97.87  & \textbf{99.94}  & 98.39  & 98.98  & 99.81  \\
8  & \textbf{96.95}  & 93.92  & 94.83  & 95.97  & 96.62  & 92.13  & 94.45  & 96.53  \\
9  & 98.44  & 98.82  & \textbf{98.91}  & 96.20  & 96.24  & 97.78  & 98.48  & 98.76  \\
\hline
OA & 97.80  & 96.90  & 92.84  & 98.48  & 97.01  & 98.17  & 98.98  & \textbf{99.18}  \\
AA & 98.50  & 97.63  & 95.03  & 98.48  & 97.39  & 98.14  & 98.88  & \textbf{99.15}  \\
K  & 97.12  & 95.94  & 90.80  & 98.01  & 96.09  & 97.61  & 98.66  & \textbf{98.93}  \\
\hline
\end{tabular}
}
\end{threeparttable}
\end{table}

\begin{table}[!t]
\setlength{\abovecaptionskip}{0.cm}
\setlength{\belowcaptionskip}{-0.cm}
\renewcommand{\arraystretch}{1.1}
\caption{Comparison Experimental Results on WHU-Hi-Hanchuan Using 150 Training Samples Each Class}
\label{tab:hanchuan_150}
\centering
\begin{threeparttable}
\resizebox{\linewidth}{!}{
\begin{tabular}{c|ccccccc|c}
\hline
Models & 3D-CNN & LWNet & SpectralFormer & GraphGST & SSFTT  & MHCFormer  & MASSFormer & Tri-Former \\
\hline
1  & 92.21  & 89.85  & 87.06  & 98.32  & 93.14  & 97.21  &\textbf{98.55}  & 97.82  \\
2  & 95.84  & 95.09  & 84.90  & 95.34  & 82.14  & 92.92  & \textbf{98.35}  & 98.13  \\
3  & 93.01  & 93.94  & 91.71  & \textbf{99.96}  & 94.83  & 98.92  & \textbf{99.99}  & 99.72  \\
4  & 96.75  & 95.87  & 96.90  & \textbf{99.85}  & 97.71  & 99.44  & 99.65  & 99.62  \\
5  & \textbf{100.00} & \textbf{100.00} & \textbf{100.00} & \textbf{100.00}  & 99.43  & \textbf{100.00} & \textbf{100.00} & \textbf{100.00} \\
6  & 90.12  & 87.66  & 84.95  & 98.29  & 91.95  & 93.63  & \textbf{99.61}  & 97.26  \\
7  & 96.96  & 96.44  & 96.68  & 99.84  & 96.84  & 98.73  & 99.65  & \textbf{99.91}  \\
8  & 90.50  & 85.05  & 81.43  & 95.67  & 86.37  & 93.99  & 95.87  & \textbf{98.60}  \\
9  & 92.64  & 86.40  & 91.50  & 95.45  & 93.23  & 90.02  & 97.69  & \textbf{97.86}  \\
10 & 92.92  & 89.92  & 96.69  & 97.69  & 99.22  & 98.76  & 99.23  & \textbf{99.44}  \\
11 & 96.92  & 94.31  & 95.63  & 97.03  & 95.87  & 97.70  & 96.97  & \textbf{99.45}  \\
12 & 98.53  & 97.00  & 99.18  & \textbf{100.00}  & 96.32  & 99.86  & \textbf{100.00} & 99.83  \\
13 & 89.46  & 85.09  & 88.04  & 92.17  & 82.52  & 84.72  & 92.65  & \textbf{95.61}  \\
14 & 96.95  & 94.20  & 87.54  & 97.61  & 93.26  & 95.06  & 97.27  & \textbf{97.72}  \\
15 & \textbf{100.00} & 98.07  & 99.47  & \textbf{100.00}  & 93.81  & \textbf{100.00} & 99.90  & \textbf{100.00} \\
16 & 95.70  & 96.28  & 98.09  & 98.98  & 99.09  & 98.04  & \textbf{99.48}  & 99.07  \\
\hline
OA & 94.38  & 92.76  & 91.76  & 97.75  & 93.81  & 96.50  & 98.37  & \textbf{98.56}  \\
AA & 94.91  & 92.82  & 92.49  & 97.89  & 93.48  & 96.24  & 98.43  & \textbf{98.75}  \\
K  & 93.45  & 91.56  & 90.41  & 97.37  & 92.76  & 95.93  & 98.09  & \textbf{98.31}  \\
\hline
\end{tabular}
}
\end{threeparttable}
\end{table}

\begin{table}[!t]
\setlength{\abovecaptionskip}{0.cm}
\setlength{\belowcaptionskip}{-0.cm}
\renewcommand{\arraystretch}{1.1}
\caption{Comparison Experimental Results on WHU-Hi-Honghu Using 150 Training Samples Each Class}
\label{tab:honghu_150}
\centering
\begin{threeparttable}
\resizebox{\linewidth}{!}{
\begin{tabular}{c|ccccccc|c}
\hline
Models & 3D-CNN & LWNet & SpectralFormer & GraphGST & SSFTT  & MHCFormer  & MASSFormer & Tri-Former \\
\hline
1  & 94.46  & 91.75  & 95.98  & \textbf{99.05}  & 98.79  & 96.87  & 96.31  & 97.34  \\
2  & 97.29  & 97.95  & 94.68  & 98.13  & 95.54  & 95.12  & 97.44  & \textbf{98.30}  \\
3  & 90.55  & 91.82  & 90.55  & 95.60  & 92.11  & 94.30  & 92.53  & \textbf{95.99}  \\
4  & 96.71  & 95.02  & 96.09  & 99.03  & 96.15  & 98.97  & 97.50  & \textbf{99.30}  \\
5  & 98.52  & 98.38  & 95.59  & \textbf{100.00}  & 97.82  & 99.77  & 99.70  & 99.51  \\
6  & 94.77  & 91.94  & 91.24  & \textbf{98.95}  & 96.52  & 98.51  & 97.11  & 98.62  \\
7  & 79.01  & 70.19  & 77.25  & \textbf{95.02}  & 87.22  & 87.66  & 91.38  & 93.04  \\
8  & 92.90  & 96.62  & 76.57  & \textbf{100.00}  & 96.41  & 99.72  & 99.23  & \textbf{100.00}  \\
9  & 97.82  & 94.26  & 98.76  & 98.97  & 98.65  & 98.76  & 96.28  & \textbf{99.22}  \\
10 & 94.79  & 94.89  & 80.35  & 98.14  & 88.36  & 98.50  & 98.46  &\textbf{99.26}  \\
11 & 92.31  & 83.66  & 80.49  & 96.30  & 93.33  & 96.70  & 96.02  & \textbf{96.84}  \\
12 & 90.57  & 85.81  & 83.72  & \textbf{98.86}  & 85.29  & 95.35  & 96.71  & 97.40  \\
13 & 91.39  & 83.51  & 77.12  & 95.06  & 87.41  & 93.62  & \textbf{98.08}  & 96.77  \\
14 & 96.14  & 97.21  & 89.52  & \textbf{99.33}  & 91.35  & 97.63  & 97.00  & 99.18  \\
15 & \textbf{100.00}  & 99.41  & 98.60  & 99.88  & 99.30  & 99.88  & \textbf{100.00}  & \textbf{100.00}  \\
16 & 99.76  & 99.25  & 95.92  & 99.80  & 94.05  & 99.65  & 98.02  & \textbf{100.00}  \\
17 & 98.85  & 98.64  & 97.64  & 99.79  & 98.50  & \textbf{100.00}  & 99.72  & 99.51  \\
18 & 98.79  & 97.75  & 96.58  & \textbf{99.05}  & \textbf{99.05}  & 98.89  & 98.04  & 98.73  \\
19 & 92.78  & 89.22  & 92.72  & 98.91  & 95.59  & 96.98  & 98.49  &\textbf{98.91}  \\
20 & 97.54  & 99.70  & 97.79  & 99.22  & 97.99  & \textbf{99.79}  & 98.65  & 99.10  \\
21 & 99.75  & 99.92  & 98.04  & \textbf{100.00}  & 98.47  & \textbf{100.00}  & \textbf{100.00}  & \textbf{100.00}  \\
22 & 98.43  & 99.41  & 96.73  & \textbf{100.00}  & 98.77  & \textbf{100.00}  & 99.25  & 99.77  \\
\hline
OA & 94.37  & 91.83  & 91.40  & 98.29  & 94.46  & 97.35  & 96.44  & \textbf{98.31}  \\
AA & 95.14  & 93.47  & 91.00  & \textbf{98.59}  & 94.85  & 97.58  & 97.29  & 98.49  \\
K	&92.92 	&89.75 	&89.24 	&97.84 	&93.04 	&96.65	&95.51	&\textbf{97.87} \\

\hline
\end{tabular}
}
\end{threeparttable}
\end{table}

Fig.~\ref{fig:indian} shows the corresponding visual results. At the top, a color-composite image is presented, accompanied by color codes that represent different types of ground objects. The bottom-right corner features a magnified view, aimed at providing a clearer comparison of the classification performance of different methods. From Fig.~\ref{fig:indian}, we can see that all methods exhibit varying degrees of misclassification. Our proposed Tri-Former demonstrates the least amount of misclassification, approaching the ground truth closely. 

Similar trends are also demonstrated in the experiments conducted on the Pavia University and Houston University datasets. When evaluated on the Pavia University dataset, the proposed Tri-Former algorithm surpasses the state-of-the-art method SSFTT, achieving remarkable improvements in key evaluation metrics. Specifically, we observe substantial enhancements in Overall Accuracy (OA), Average Accuracy (AA), and Kappa (K) coefficient by 1.85\%, 1.71\%, and 2.45\%, respectively. On the Houston University dataset, where the potential for performance improvement is considerably limited, our Tri-Former still manages to achieve the best performance among all methods compared. Fig.~\ref{fig:paviau} and Fig.~\ref{fig:hu} are the corresponding visual comparison results. On these two datasets, especially on the Pavia University dataset, the proposed Tri-Former produces classification result maps that closely resemble the groundtruth.

Across all three WHU-Hi datasets, the proposed Tri-Former consistently outperforms conventional models like 3D-CNN, LWNet, and even newer transformer-based models such as SpectralFormer, GraphGST, and SSFTT. This can be attributed to Tri-Former's ability to effectively capture both spatial and spectral features while handling the unique challenges of hyperspectral imaging data. The results from Table~\ref{tab:longkou_150} to Table~\ref{tab:honghu_150} substantiate the argument that incorporating hybrid transformer structures, as demonstrated in Tri-Former, is a highly effective approach for HSI classification.

\subsection{Inference speed analysis on Tri-Former}

To comprehensively evaluate the complexity and the inference speed of the proposed Tri-Former model, we conducted a thorough comparison between Tri-Former and two other recent, representative transformer-based HSI classification methods: SpectralFormer and SSFTT.

\begin{figure}
	\begin{center}
		\includegraphics[width=0.7\linewidth]{ 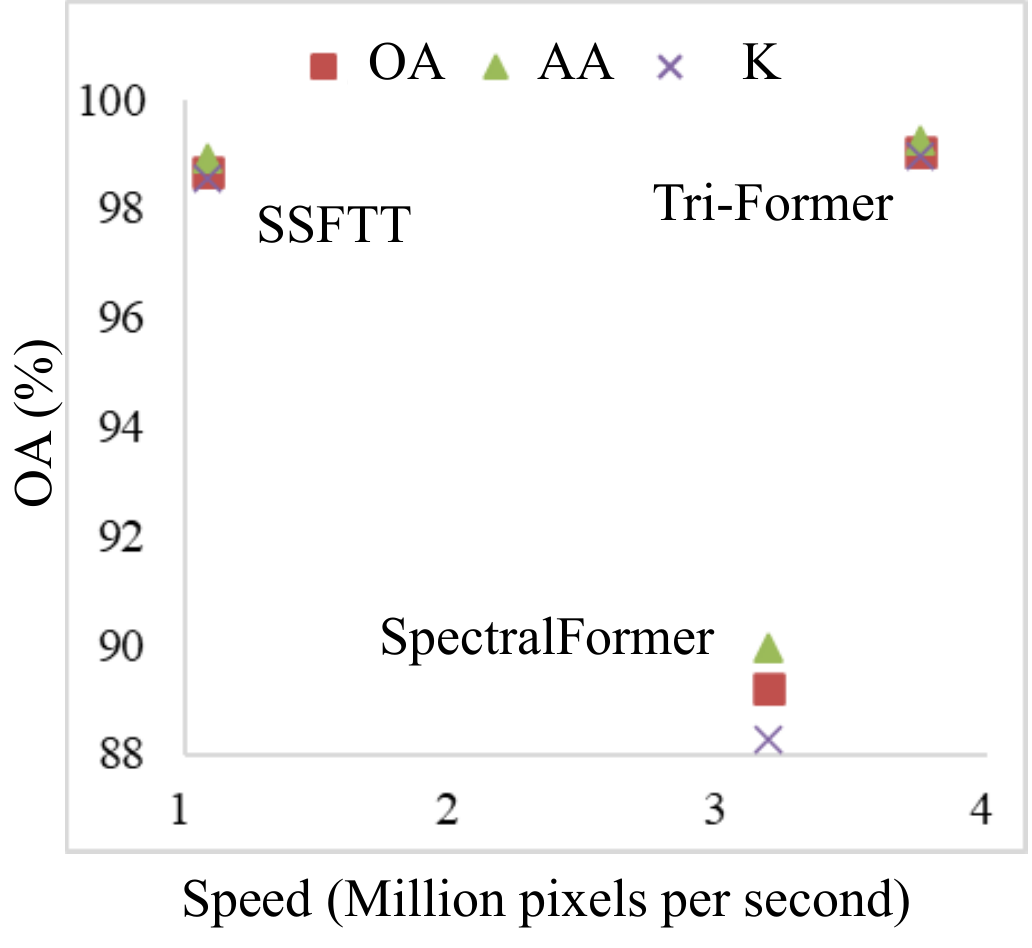}
	\end{center}
	\vspace{-0.3cm}
	\caption{Comparison results on classification accuracy and inference speed.}
	\label{fig:speed}
\end{figure}

In this section, we ran each model 10 times on the same dataset and averaged the number of pixels processed per second. The experimental results are shown in Fig.~\ref{fig:speed}, which demonstrate that the proposed Tri-Former enjoy clear advantages in both classification accuracy and inference speed. Compared with SSFTT, Tri-Former achieves 3$\times$ faster inference speed, while maintaining advantage in classification accuracy.  On the Houston University dataset, Tri-Former improves classification accuracy by 10\% compared to SpectralFormer while also achieving about a 20\% increase in inference speed.

\begin{figure}[t]
	\begin{center}
		\includegraphics[width=0.95 \linewidth]{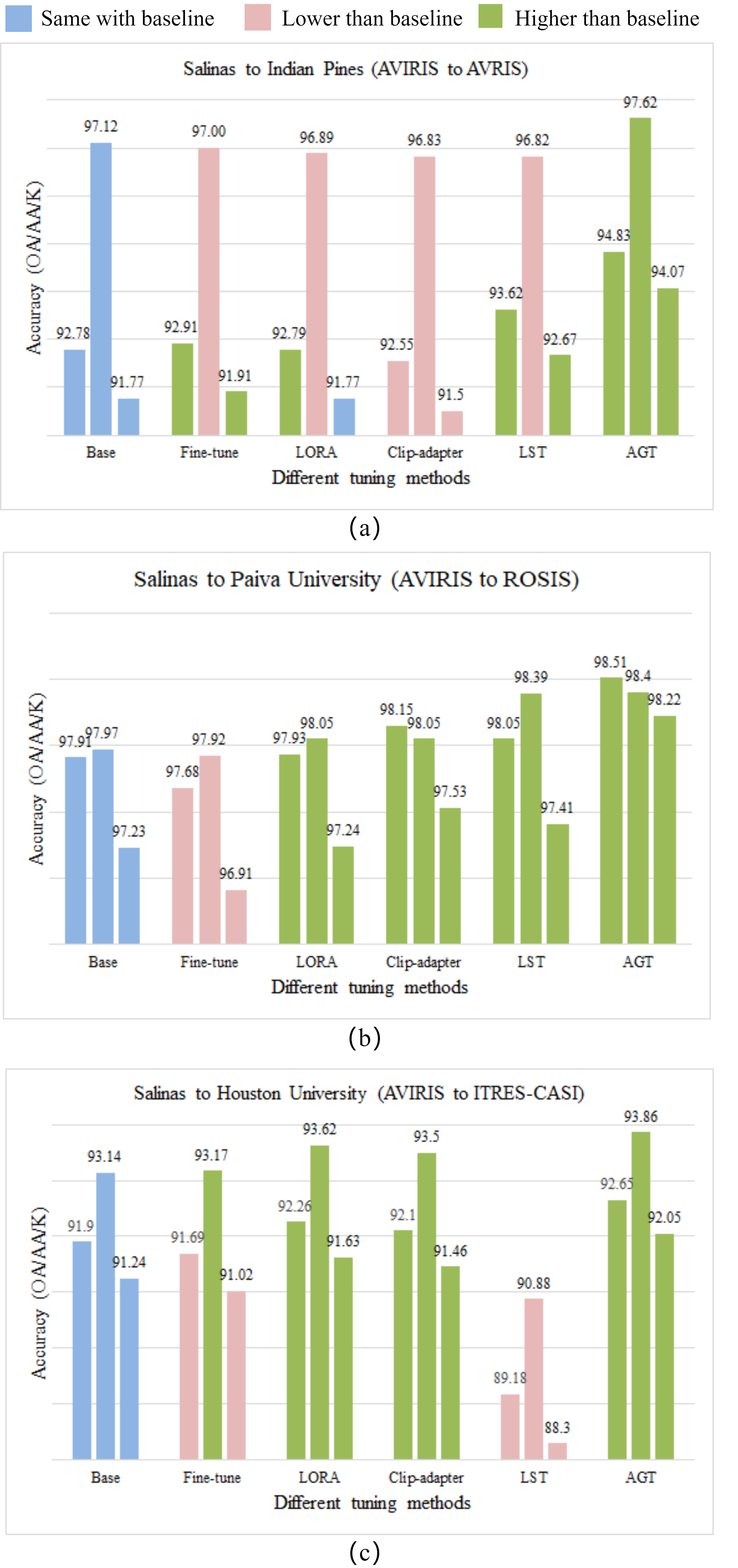}
	\end{center}
	\vspace{-0.3cm}
	\caption{Comparison of different tuning methods.}
	\label{fig:TF_comparison}
 \vspace{-0.3cm}
\end{figure}

\begin{figure}
	\begin{center}
		\includegraphics[width=1.0 \linewidth]{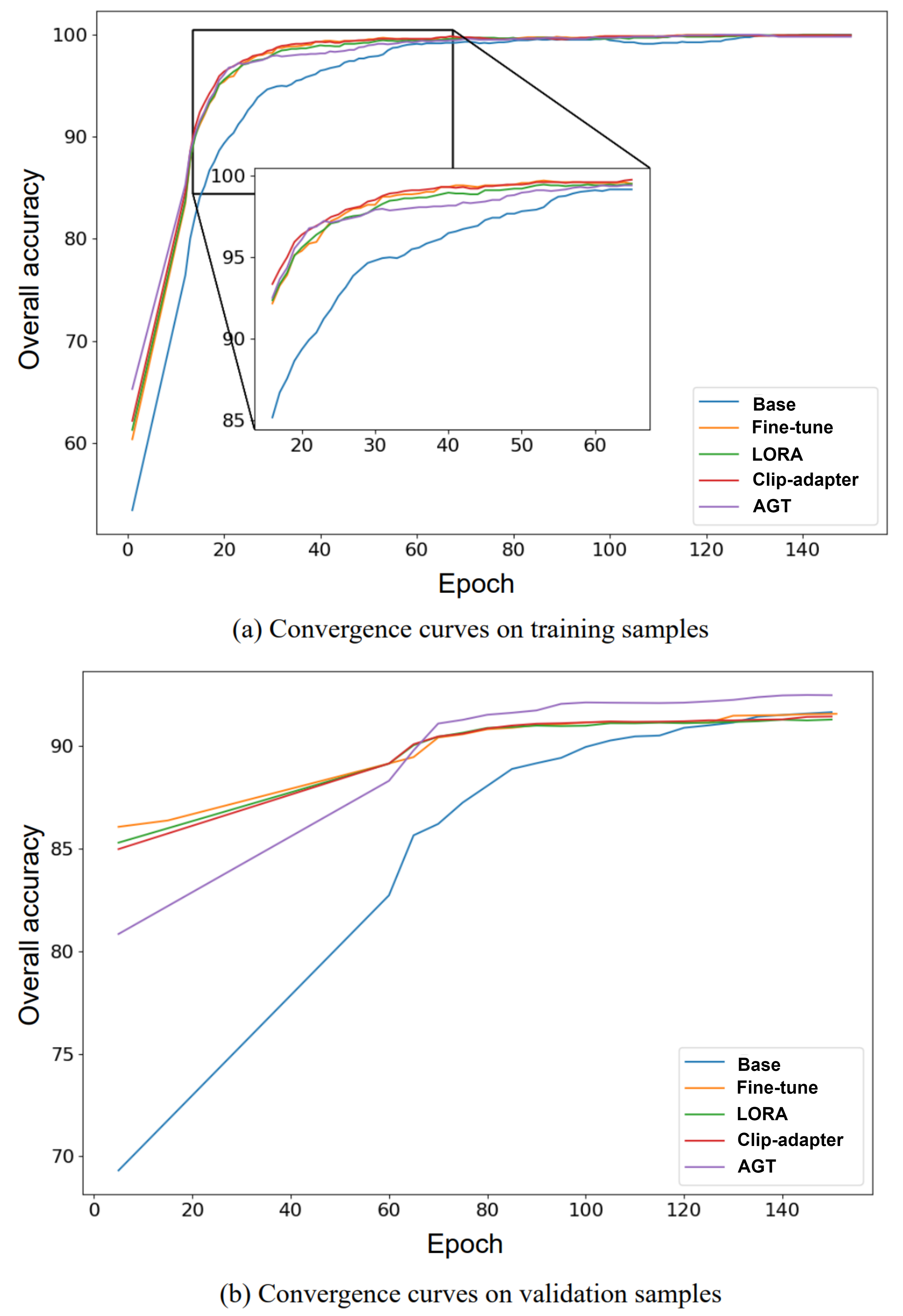}
	\end{center}
	\vspace{-0.3cm}
	\caption{Convergence curves of different architecture tuning strategies.}
	\label{fig:convergence_curves}
\end{figure}

\begin{figure*}
	\begin{center}
		\includegraphics[width=0.8 \linewidth]{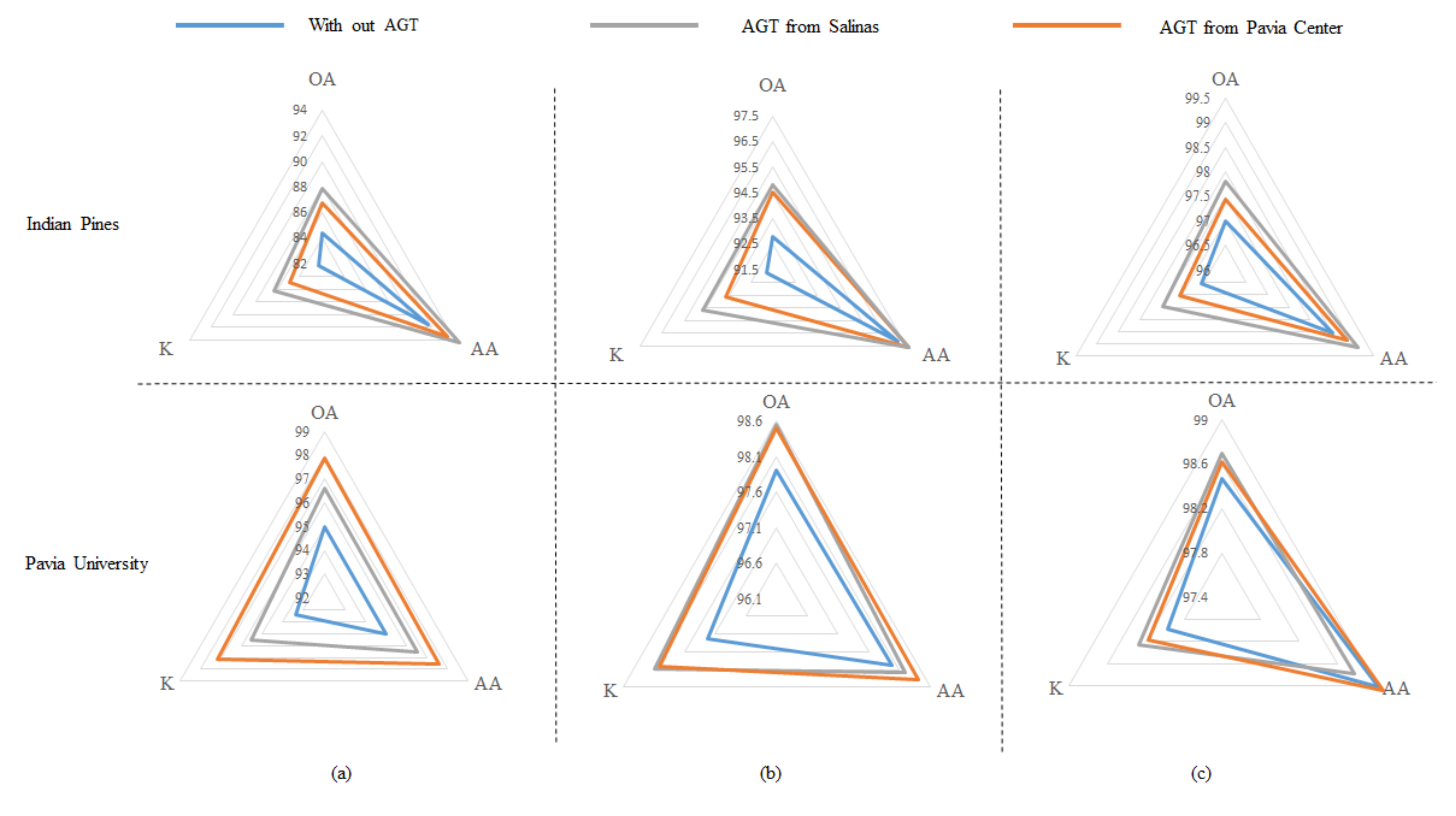}
	\end{center}
	\vspace{-0.6cm}
	\caption{Experimental results on homogeneous and heterogeneous HSI datasets. (a) 25 training samples per class; (b) 50 training samples per class; (c) 75 training samples per class.}
	\label{fig:tf_lidar}
\end{figure*}

\subsection{Ablation study on Tri-Former}

\begin{table}[t]
\scriptsize
\renewcommand\arraystretch{1.0}
\begin{center}
\caption{Ablation study}
\label{tab:ablation}
\begin{tabular}{ccc|ccc}
\hline
\multicolumn{3}{c|}{Components} & \multicolumn{3}{c}{Performance (\%)} \\
Spectral module &  Spatial module & 3D Conv     & OA & AA & K\\
\hline
 \checkmark     &   \checkmark    &    -         & 97.26   & 97.31   & 96.83  \\
                &   \checkmark    &  \checkmark  & 98.06   & 97.97   & 97.76  \\
 \checkmark     &                 &  \checkmark  & 97.26   & 97.05   & 96.84  \\
 \checkmark     &    \checkmark   &  \checkmark  & 98.41   & 97.87   & 98.17  \\
\hline
\end{tabular}
\end{center}
\end{table}

To evaluate the function of each component in the proposed Tri-Former module, we conducted ablation study experiments on Indian Pines with 150 training samples each class. To ensure that the performance degradation is not due to the reduction in model parameters, the removed parts in the ablation experiment will be replaced with pointwise convolution or 3D convolution of equivalent size. The corresponding results are listed in Table \ref{tab:ablation}. From the experimental results, it can be concluded that each component of the design is necessary. Among them, the 3D convolution module has the greatest impact, which we attribute to its ability to simultaneously extract spatial-spectral features.

\subsection{Comparison between AGT and state-of-the-art methods}

For evaluating the effectiveness of the proposed AGT strategy, we conducted a comprehensive comparison with several representative architecture tuning methods. The comparison methods involved in the evaluation are as follows: Fine-tune (adopted in LWNet~\cite{2019Hyperspectral} and Two-CNN~\cite{yang2017learning} for ConvNet methods), LORA~\cite{hu2022lora}, Clip-adapter~\cite{gao2021clip} and LST~\cite{sung2022lst} (for transformer methods). 

In this experiment, the basic model is first pretrained on the Salinas dataset acquired through the AVRIS sensor. In this process, 50 labeled pixels are randomly extracted from each class to build a pretraining set. Then, the model is fine-tuned with various strategies on three distinct target datasets captured by different sensors: AVRIS, ROSIS and ITRES-CASI. Within the tuning stage, each class is provided with 50 training samples. The experimental results are consolidated in Fig.~\ref{fig:TF_comparison}, where the x-axis denotes the employed tuning methods and the y-axis depicts the evaluation metric. According to the comparison, we can draw four conclusions:

\begin{itemize}

\item \textbf{Conventional fine tuning strategy tailored for ConvNets proves ineffective on the cross-sensor HSI data.} As shown in Fig.~\ref{fig:TF_comparison}, Fine-tune strategy not only fails to yield any improvement but also negatively affects the final results. We hypothesize this phenomenon can be attributed to the fundamental differences between the information learned by ConvNets and transformer architectures. ConvNets learn a set of filter parameters from training data. The process of feature extraction in ConvNets is static. Differing from this, transformer architectures learn relationships between pixels, capturing global dependencies in the data. The transformer is a data-driven model with a dynamic feature extraction process, making it more sensitive to differences between pretrain dataset and target dataset. Unfortunately, there are obvious differences among cross-sensor HSI datasets, which consequently hinder effectiveness of fine tuning.

\item \textbf{Tuning strategies proposed for transformer architecture have a certain role on cross-sensor HSI data.} LORA and Clip-adaptor are specifically designed for transformer-based algorithms, and they do achieve higher classification accuracy compared to fine tuning methods tailored for CNNs. However, despite their overall performance advantage, these two fine tuning methods still exhibit negative effects in certain experiments. We speculate the reason is that the pretrained basic model is heavily influenced during the tuning stage. When the adaptor introduces significant changes to the basic model during fine tuning, the fine-tuned model may lose some of the informative features learned during pre-training. As a result, the performance of LORA and Clip-adapter could be negatively impacted in certain scenarios, particularly when there are a lot of differences between the pre-training and fine tuning datasets. Significant updating of the basic model's parameters may disrupt the underlying spectral-spatial patterns captured during pre-training, leading to a drop in classification accuracy.

\item \textbf{Freezing the basic model is not always effective.} As shown in Fig.~\ref{fig:TF_comparison}, LST does not perform well in \textit{Salinas$\rightarrow$Houston University} experiment, whereas it brings in certain improvements in the other two experiments. We speculate that this issue arises from the LST strategy, which does not update the gradients of the basic model. While this approach preserves the features learned during pre-training, it is not always beneficial when there are significant differences in data structure, such as the number of bands. In such cases, the pre-trained knowledge may conflict with the target dataset. These conflicting representations are not adjusted during LST fine tuning, leading to performance drop.

\item \textbf{Proposed AGT works well.} Experimental results have shown that information selection and gradient updates during fine tuning are crucial. Benefiting from the attention-gated block and cold-hot gradient update strategy, AGT ensures that the fine tuning process adapts the model to the target dataset, while preserving the generic spectral-spatial representations learned during pre-training. This approach contributes to its superior performance in various experimental scenarios, even when faced with cross-sensor data.

\item \textbf{Architecture tuning does have an impact on training effectiveness.} To conduct a comprehensive analysis of different tuning strategies, we have plotted their respective convergence curves. As shown in Fig~\ref{fig:convergence_curves}, compared with baseline, each tuning strategy exhibits a distinct convergence curve. Specifically, 
each tuning model achieves very high accuracy at beginning. The accuracy of training the base model (\textcolor{blue}{in blue line}) gradually catches up with the accuracy curves of Fine-tune, Clip-adapter, and LORA. However, it still lags behind the convergence curve of AGT. We attribute this phenomenon mainly to the following reason. At the beginning of training, the baseline model starts with random initialization and lacks any feature extraction capability. Meanwhile, the tuned models have already learned feature extraction capabilities from heterogeneous data, allowing them to exhibit some effectiveness from the outset. During the training process, the baseline model benefits from the optimization on the new data, while tuned models may suffer from the conflicts gradually arising from the heterogeneous data. This dynamic change results in the baseline catching up with, and in some cases, surpassing the tuned models.

\end{itemize}

\subsection{Experiments on homogeneous and heterogeneous HSI datasets}

To further validate the effectiveness of the proposed method and investigate the influence of homogeneity and heterogeneity on tuning experiments, we conducted two sets of experiments on the Indian Pines, Salinas, Pavia University, and Pavia Center datasets. The first two datasets, Indian Pines and Salinas, are homogeneous and captured by AVIRIS, while the last two, Pavia University and Pavia Center, are homogeneous and captured by ROSIS. In the first set of experiments, the basic model was pretrained on the Salinas dataset and fine-tuned on the Indian Pines and Pavia University datasets under three configurations: 25, 50, and 75 training samples per class. The experimental results depicted in the radar charts in Fig.~\ref{fig:tf_lidar} show that:

\begin{itemize}
    \item \textbf{AGT boosts performance across all the experimental settings.} Whether in a homogeneous or heterogeneous scenario, employing AGT consistently leads to a significant improvement in accuracy, which once again proves the effectiveness of our attention-gated strategy.

    \item \textbf{Transferring between homogeneous datasets offers more advantages than transferring between heterogeneous datasets.} Even Pavia Center dataset contains only ten classes and has less sample diversity compared to Salinas dataset with 16 classes, the advantages of \textit{Pavia Center$\rightarrow$Pavia University} tuning are more obvious than \textit{Salinas$\rightarrow$Pavia University} tuning. This phenomenon shows that when fine tuning transformer models, the similarity between pretrain and target datasets is likely to be important than the diversity of the pretrain dataset itself. This differs from previous experiences with fine tuning ConvNets. Researchers~\cite{2019Hyperspectral} found that for ConvNets, the diversity of the pretrain data is more important than its homogeneity. We speculate this is because transformers learn dependencies between pixels in the dataset, while ConvNets learn static filter parameters from the data. Therefore, fine tuning transformers requires higher homogeneity between pretrain and fine tuning data.
    
\end{itemize}

\begin{table}  
\renewcommand{\arraystretch}{1.1}
\setlength{\tabcolsep}{2.5pt}
  \caption{Experimental Results of Cross Sensors Transfer Learning from RGB Data}
  \centering
    \begin{tabular}{c|ccc|ccc|ccc}
    \hline
    Dataset & \multicolumn{3}{c|}{Indian pines} & \multicolumn{3}{c|}{PaviaU} & \multicolumn{3}{c}{HoustonU} \\
    \hline
    Sensor & \multicolumn{3}{c|}{RGB$\rightarrow$ AVIRIS} & \multicolumn{3}{c|}{RGB$\rightarrow$ ROSIS} & \multicolumn{3}{c}{RGB $\rightarrow$ ITRES-CASI} \\
    
    Method & Base   & LST   & AGT   & Base   & LST    & AGT   & Base & LST    & AGT \\
    \hline
    OA    &   84.36 & 70.14 & \textbf{86.14} & 95.00  & 81.05  & \textbf{95.08} & 83.00 & 71.00 & \textbf{85.14} \\    
    AA    &   91.64 & 83.33 & \textbf{93.85} & 95.03  & 81.21  & \textbf{95.40} & 84.34 & 75.06 & \textbf{87.17} \\    
    K     &   82.28 & 66.46 & \textbf{84.35} & 93.40  & 75.43  & \textbf{93.61} & 81.64 & 68.72 & \textbf{83.96} \\
    \hline
    \end{tabular}%
  \label{tab:tf_rgb}%
\end{table}%

\subsection{Using AGT as a bridge between RGB datasets and HSI datasets}
RGB data is abundant and highly diverse. The barrier between RGB and HSI datasets is that they are cross-modality datasets. It is also important to note that both hyperspectral and RGB images are acquired by optical sensors, and thus, share similar spatial characteristics due to the common imaging mechanisms of light reflection, refraction, and diffraction. Moreover, most hyperspectral datasets inherently contain the RGB bands, and many hyperspectral sensors exhibit overlapping spectral ranges in these bands. These commonalities ensure that the spectral and spatial features of optical data are highly transferable. In contrast, SAR images, which are generated by active microwave radar systems, rely on scattering mechanisms that yield fundamentally different image characteristics. As a result, effective knowledge transfer between SAR and optical datasets maybe much more challenging.  

In this section, we combine HSI reconstruction method and the proposed AGT to build a bridge to cross this barrier. More specifically, a pretrained HSI reconstruction model~\cite{li2022drcr} is applied to convert 3-channel RGB image to 32-channel pseudo-HSIs. Then the pseudo-HSIs are used to pretrain the basic model. The RGB dataset used here is CIFAR\footnote{\url{https://www.cs.toronto.edu/~kriz/cifar.html}}. We randomly extract 25 samples from each category as tuning samples and take the rest as the testing samples. Experimental results are listed in Table~\ref{tab:tf_rgb}. 

Interestingly, in this setting, AGT achieves results that are as good as, or even superior to, those obtained with homogenous data. Cross-modality experiments can be viewed as a further extension of cross-sensor experiments. At first glance, this experimental result seems contradictory to the previous results. We suspect that the main reason for this phenomenon is the strong sample diversity of the CIFAR dataset, which offsets the effects of heterogeneous data to some extent. This is essential for remote sensing applications with expensive data collection costs.

Just as mentioned in our method, the proposed AGT serves as a bridge to address the modal differences between RGB datasets and HSI datasets. This is crucial, particularly considering the presence of datasets with even greater sample diversity in the realm of RGB datasets. The bridge allows us to make use of these datasets. Additionally, to enhance the HSI classification performance, larger models are required. Larger labeled datasets are necessary for training these larger models. AGT provides a way to effectively leverage existing RGB datasets to enhance the accuracy of HSI classification.

\subsection{Analysis on attention gate}

\begin{figure}
	\begin{center}
		\includegraphics[width=1.0 \linewidth]{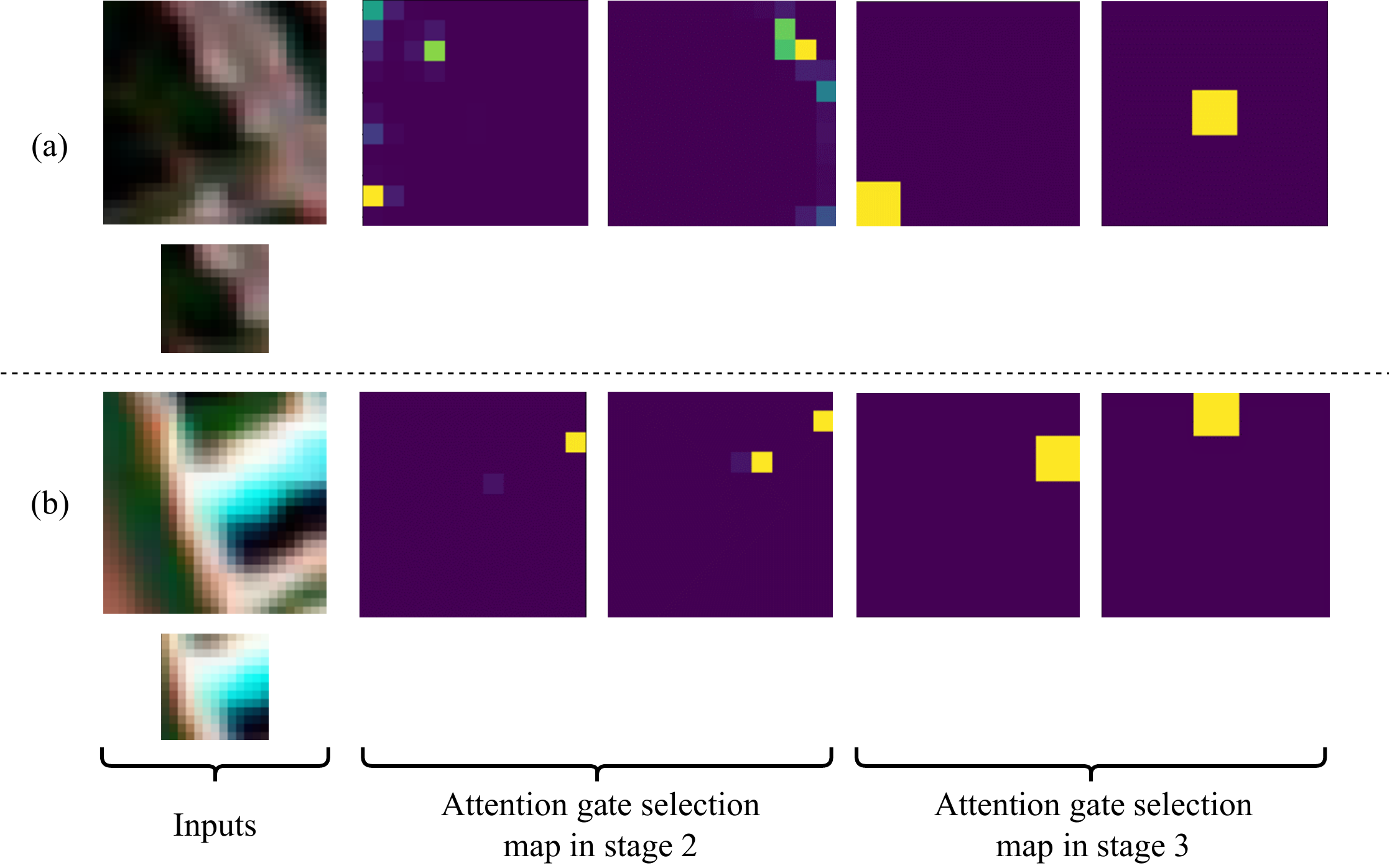}
	\end{center}
	\vspace{-0.3cm}
	\caption{Visual results of gate attention selection process. (a) and (b) show two different samples. }
	\label{fig:gate_attention}
\end{figure}

In this section, we investigate whether the proposed attention gate can select relevant features from the basic model to enhance the features in the auxiliary branch. Specifically, following the inference process, we feed a large patch into the basic model and crop a smaller patch from its center to input into the auxiliary branch. The corresponding visual results are presented in Fig.~\ref{fig:gate_attention}.

\begin{table}[t]
\scriptsize
\renewcommand\arraystretch{1.0}
\setlength\tabcolsep{8pt}
\begin{center}
\caption{Ablation study on the proposed attention gate}
\label{tab:ablation_gate}
\begin{tabular}{ccc|ccc}
\hline
Methods  &  Dataset  &  Number of samples  & OA  & AA  &  K \\
\hline
Addition   &  Salinas to    &    50      & 92.71 &  96.71  &  91.68 \\
AGT      &  Indian Pines  &    50      & 94.83 &  97.62  &  94.07 \\
\hline
\end{tabular}
\end{center}
\end{table}

The visual results demonstrate that the attention gate functions passing relevant features selectively. If we directly connected the basic model and auxiliary branch, all features in the basic model  be equally selected. Alternatively, if the proposed attention mechanism could not guide information flow from the basic model  to the auxiliary branch, the selection map would appear random. However, Fig.~\ref{fig:gate_attention} shows that features from the basic model highly relevant to those in the auxiliary branch are selectively chosen. This observation underscores the effectiveness of the attention gate mechanism in fine tuning HSI classification methods. As shown in Table \ref{tab:ablation_gate}, if we replace the proposed attention gate architecture with a simple addition operation, the performance drops significantly, which also illustrates that the proposed attention gate is effective.

\section{Conclusion}

In this paper, to effectively diminish the demand for an extensive quantity of training samples, we propose the attention gated tuning (AGT) strategy. Acting as a bridge, AGT enables us to capitalize on both pre-existing labeled HSI datasets and even RGB datasets, thereby enhancing the performance on new HSI datasets with constrained sample sizes. Furthermore, we have presented the triplet-structured transformer (Tri-Former) model for HSI classification. By introducing spectral-spatial parallel block, Tri-Former optimizes computational efficiency while enhancing model stability. Extensive experiments across diverse HSI datasets and sensors verified the proposed Tri-Former. Homologous, heterologous, and cross-modality tuning experiments are conducted to validate the performance of the proposed AGT. We hope this research can inspire the designs of fine tuning methods in HSI classification and guide the selection of pre-training datasets.

\appendices



\ifCLASSOPTIONcaptionsoff
  \newpage
\fi




\bibliographystyle{IEEEtran}
\bibliography{reference}

\end{document}